\documentclass[a4paper,fleqn]{cas-dc}

\usepackage[numbers]{natbib}
\usepackage{tabularx,booktabs,graphicx,amsmath}
\usepackage{balance,makecell,multirow,array,longtable}
\usepackage{tikz}
\usepackage{amssymb,amsmath}
\usepackage{pgfplots}
\pgfplotsset{compat=1.18}
\usetikzlibrary{plotmarks}
\usetikzlibrary{matrix, trees, positioning, shapes, arrows.meta, calc}
\usepackage{hyperref}
\usetikzlibrary{shapes.geometric}
\usetikzlibrary{pgfplots.groupplots}
\usepackage{placeins}
\usepackage[dvipsnames]{xcolor}
\usepackage{rotating}
\usepackage{adjustbox}
\usepackage{float}
\usepackage{subcaption}
\usepackage[percent]{overpic}
\newcolumntype{Y}{>{\centering\arraybackslash}X}
\graphicspath{{figures/}}

\begin{document}

\let\WriteBookmarks\relax
\renewcommand{\topfraction}{0.9}
\renewcommand{\bottomfraction}{0.8}
\renewcommand{\textfraction}{0.05}
\renewcommand{\floatpagefraction}{0.75}
\renewcommand{\dbltopfraction}{0.9}
\renewcommand{\dblfloatpagefraction}{0.75}
\setcounter{topnumber}{4}
\setcounter{bottomnumber}{2}
\setcounter{totalnumber}{6}
\setcounter{dbltopnumber}{3}

\shorttitle{Fake Review Detection: A Survey}
\shortauthors{Yang et al.}

\title[mode=title]{A Survey on Fake Review Detection: From Pre-trained Language Models to Large Language Models}


\author[1]{Fanji Yang}
\fnmark[1]
\ead{FanjiYang@mail.gufe.edu.cn}
\credit{Data curation, Investigation, Writing - original draft}

\author[2]{Huiyao Chen}
\fnmark[1]
\ead{chenhy1018@gmail.com}
\credit{Data curation, Investigation, Writing - original draft}

\author[1]{Xi Yu}
\cormark[1]
\ead{yuxi@mail.gufe.edu.cn}
\credit{Supervision, Writing - review \& editing}

\author[2]{Meishan Zhang}
\ead{mason.zms@gmail.com}
\credit{Methodology, Validation}

\author[3]{Xiaohong Xiao}
\cormark[1]
\ead{xiaoxiaoh@mail.gufe.edu.cn}
\credit{Supervision}

\author[1]{Mingsen Deng}
\cormark[1]
\ead{msdeng@mail.gufe.edu.cn}
\credit{Project administration, Funding acquisition}

\fntext[1]{These authors contributed equally to this work.}

\cortext[1]{Corresponding author}


\affiliation[1]{
    organization={Guizhou University of Finance and Economics},
    city={Guiyang},
    country={China}
}

\affiliation[2]{
    organization={Harbin Institute of Technology (Shenzhen)},
    city={Shenzhen},
    country={China}
}

\affiliation[3]{
    organization={Guizhou University of Commerce},
    city={Guiyang},
    country={China}
}

\begin{abstract}
Online reviews shape consumer decisions, platform governance, and corporate reputation.
Fake reviews compromise this information channel by injecting deceptive evidence into rating systems, recommendation pipelines, and public trust mechanisms.
The rise of large language models, or LLMs, has changed the problem in two directions.
LLMs can generate fluent and context-aware deceptive reviews, while pre-trained language models, or PLMs, and LLMs also provide stronger semantic representations for detection.
This survey reviews fake review detection from an information fusion perspective, covering 211 studies published from 2018 to early 2026.
We organize existing work by evidence source and fusion level, covering review text, sentiment, rating behavior, temporal metadata, user-product graphs, multimodal content, external knowledge, and LLM-generated signals.
We trace the development from traditional machine learning and deep learning to PLM-based and LLM-based methods, and examine how different approaches combine textual, behavioral, structural, and multimodal evidence.
We also analyze reported performance trends on widely used Amazon, Yelp, and OpSpam benchmark families, while noting the limitations caused by different label construction procedures, data splits, and evaluation protocols.
Finally, we identify open problems in adversarial generation, cross-domain transfer, uncertainty-aware fusion, missing-source robustness, interpretability, and trustworthy evaluation for AI-generated deceptive content.
\end{abstract}

\begin{highlights}
\item Survey of fake review detection organized by evidence source and fusion level.
\item Analysis of text, behavior, graph, metadata, multimodal, and LLM-generated signals.
\item Taxonomy linking ML, DL, PLM, and LLM methods to fusion mechanisms.
\item Benchmark trend analysis with cautions about label and protocol differences.
\item Research agenda for uncertainty-aware, interpretable, and robust fusion.
\end{highlights}


\maketitle

\section{Introduction}
Online reviews are a major information channel in the digital economy.
More than 70\% of customers consult online reviews before making a purchase, and 63\% of consumers report that reviews directly influence their buying behavior~\cite{10.3389/fpsyg.2022.865702,FERNANDES2022103066}.
As the volume, diversity, and influence of online reviews continue to grow, so does the incentive to manipulate them.
The proliferation of fake reviews has emerged as a persistent and consequential problem across major platforms.
In 2012, TripAdvisor faced scrutiny over the authenticity of more than 50 million reviews, and Samsung was implicated in orchestrating fake negative reviews targeting competitors~\cite{10.7717/peerj-cs.2345}.
On platforms such as Amazon and Yelp, fake reviews undermine the credibility of review systems, distort market competition, and erode consumer trust.
Annual economic losses attributable to fake reviews are estimated to exceed \$152 billion~\cite{he2022market,vidanagama2020deceptive}.
Regulatory analyses further describe fake-review moderation as a large-scale governance problem that requires automated screening, manual review, and platform accountability~\cite{MartinezOtero2021FakeReviews}.
This evidence reflects both the scale of the problem and the limitations of current countermeasures.

\begin{figure}[pos=htbp]
  \centering
  \includegraphics[width=\linewidth]{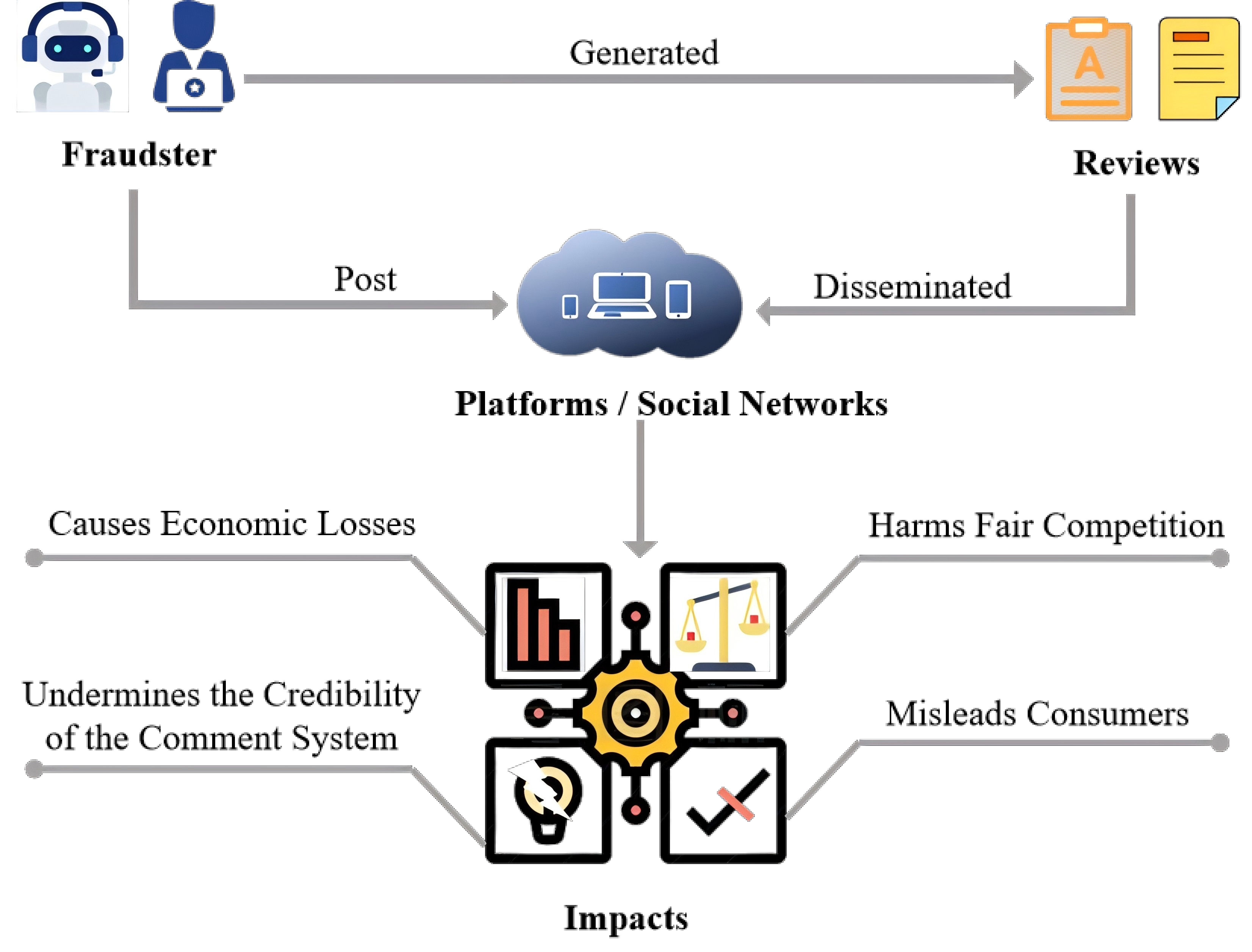}
  \caption{A conceptual framework of fake reviews, illustrating their sources, generation and dissemination mechanisms, and impacts.}
  \label{fig:conceptual_framework}
\end{figure}

Fake reviews do not arise in isolation.
As illustrated in Figure~\ref{fig:conceptual_framework}, they are produced by a range of actors, including profit-driven merchants, organized review manipulation services, and incentive-motivated users, and spread through platform recommendation algorithms and social diffusion mechanisms.
This socio-technical process amplifies the reach and impact of deceptive content, posing governance and regulatory challenges that extend well beyond simple text classification.

Research on fake review detection has evolved through several phases.
Early work relied on rule-based and traditional machine learning methods, using handcrafted features such as linguistic patterns, sentiment cues, reviewer behavior, and network structure to identify spam reviews~\cite{jindal2008opinion,li2011learning}.
The advent of deep learning enabled more expressive representation learning, with models such as convolutional neural networks, recurrent neural networks, and attention-based architectures progressively replacing manual feature engineering~\cite{mohawesh2021fake,LIU2022101865,yu2022graph}.
Fine-tuning pre-trained language models, or PLMs, such as BERT and its variants subsequently became a common strategy, yielding performance gains across multiple benchmarks~\cite{10.1145/3503162.3503169,10.1145/3497701.3497714}.

The emergence of large language models, or LLMs, has introduced a new and more complex dynamic.
On the detection side, LLMs support semantic understanding, contextual reasoning, and few-shot generalization.
On the generation side, LLMs have lowered the barrier to producing fluent deceptive content at scale.
LLM-generated fake reviews can closely mimic genuine user writing in terms of semantic consistency, sentiment expression, and linguistic style, rendering keyword-based, syntactic, and manual inspection approaches increasingly ineffective~\cite{meng2025large,zhao2025ai,amos2024consumer}.
This adversarial dynamic between generation and detection defines the central challenge of the field today.
Despite the growing body of research, existing surveys leave two gaps.
First, most surveys classify methods by model architecture or feature type, but give less attention to how heterogeneous evidence is fused across text, behavior, graphs, metadata, images, and external knowledge.
Second, the implications of LLM-generated content for source reliability, benchmark validity, and detection robustness remain underexplored~\cite{duma2025analysis}.
This survey addresses these gaps by reviewing fake review detection from 2018 to early 2026 through the joint lens of model evolution and information fusion.

The main contributions of this survey are as follows:
\begin{itemize}
    \item \textbf{Fusion-oriented taxonomy.} We organize fake review detection methods by evidence source and fusion level, covering text, sentiment, behavior, temporal metadata, graphs, multimodal content, external knowledge, and LLM-generated signals.
    \item \textbf{Model evolution with fusion mechanisms.} We review traditional machine learning, deep learning, PLM-based, and LLM-based approaches, emphasizing how each generation combines textual, behavioral, structural, and contextual evidence.
    \item \textbf{Benchmark and protocol analysis.} We compare reported trends on representative datasets while discussing how label construction, data splits, class imbalance, and evaluation metrics affect comparability.
    \item \textbf{LLM-era research agenda.} We examine LLMs as both generators of deceptive reviews and components of detection systems, and identify open problems in uncertainty-aware fusion, adversarial robustness, cross-domain adaptation, and interpretable decision-making.
\end{itemize}

\begin{figure*}[t]
    \centering
    \input{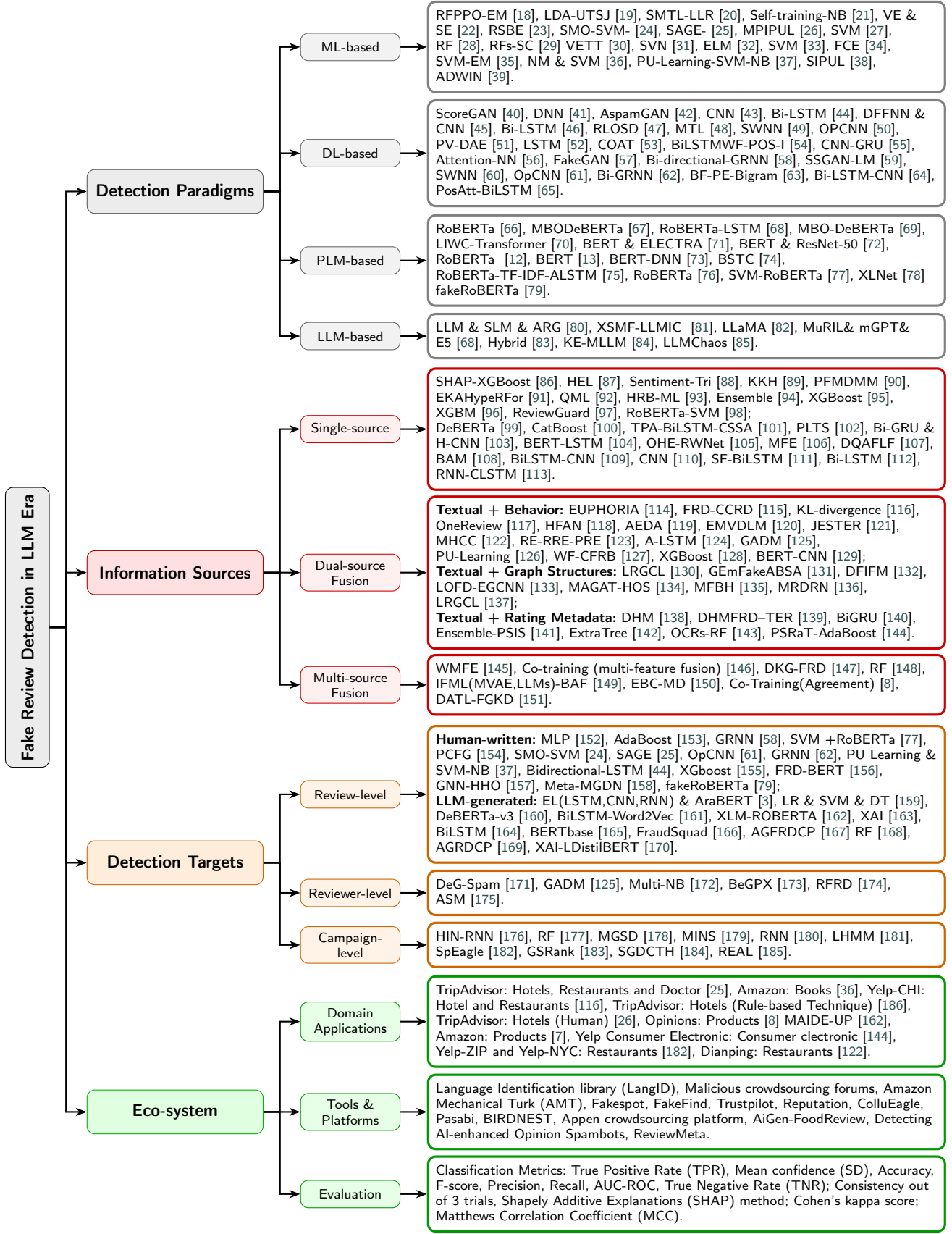}
\caption{Taxonomy of fake review detection research in the LLM era, structured by detection paradigms, information sources, detection targets, and ecosystem factors.}
    \label{fig:taxonomy}
\end{figure*}

To reflect the diversity of fake review detection research, we propose the classification system shown in Figure~\ref{fig:taxonomy}.
The remainder of this paper is organized as follows.
Section~\ref{sec:related_work} reviews recent surveys on fake review detection and summarizes their research scopes and limitations.
Section~\ref{sec:background} reviews traditional fake review detection methods.
Section~\ref{sec:PLM} focuses on PLM-based detection approaches.
Section~\ref{sec:LLM} discusses the application of LLMs in fake review detection.
Section~\ref{sec:Comparative} presents a comparative analysis of the performance evolution of state-of-the-art methods on widely used benchmark families, including Amazon, Yelp, and OpSpam.
Section~\ref{sec:challenges} addresses key challenges and future directions.
Section~\ref{sec:conclusion} concludes the paper.

\section{Related Work}
\label{sec:related_work}
Over the past decade, fake review detection has attracted attention from both academia and industry, resulting in several survey papers on this research area.
As the field has moved from feature engineering to deep representation learning, existing surveys have expanded their coverage from traditional machine learning to deep neural networks, graph-based models, and foundation models.
Table~\ref{tab:survey_comparison} summarizes representative surveys on fake review detection and compares their technical coverage and research scope.
\begin{table*}[t]
\centering
\caption{Comparison of Existing Surveys on Fake Review Detection}
\label{tab:survey_comparison}
\renewcommand{\arraystretch}{1.11}
\setlength{\tabcolsep}{5pt}

\begin{tabularx}{\textwidth}{p{3.5cm} c c c c c c X}
\toprule
\textbf{Survey} &
\textbf{Year} &
\textbf{Traditional} &
\textbf{DL} &
\textbf{PLM} &
\textbf{LLM} &
\makecell{\textbf{LLM}\\\textbf{Paradigms}} &
\textbf{Scope} \\
\midrule

~\citet{dewang2018state}
&2018
&$\checkmark$
&$\times$
&$\times$
&$\times$
&$\times$
&Feature engineering, reviewer behavior, graph-based methods, traditional ML, datasets, evaluation
\\

~\citet{vidanagama2020deceptive}
&2020
&$\checkmark$
&$\triangle$
&$\times$
&$\times$
&$\times$
&Linguistic, behavioral and metadata features, traditional ML, early DL methods, datasets and evaluation
\\

~\citet{li2020fake}
&2020
&$\checkmark$
&$\triangle$
&$\times$
&$\times$
&$\times$
&Systematic review of machine learning-based fake review detection, feature engineering, datasets, evaluation metrics, hospitality applications
\\

~\citet{mohawesh2021fake}
&2021
&$\checkmark$
&$\checkmark$
&$\times$
&$\times$
&$\times$
&Feature engineering, traditional ML, deep learning, datasets, evaluation, challenges
\\

~\citet{paul2021fake}
&2021
&$\checkmark$
&$\checkmark$
&$\triangle$
&$\times$
&$\times$
&Systematic review of feature engineering, reviewer behavior, graph-based methods, ML, DL, benchmark datasets, evaluation metrics, research challenges
\\

~\citet{mewada2022research}
&2022
&$\checkmark$
&$\checkmark$
&$\times$
&$\times$
&$\times$
&Feature engineering, traditional ML, deep learning, datasets, evaluation metrics
\\

~\citet{yu2022graph}
&2022
&$\times$
&$\checkmark$
&$\times$
&$\times$
&$\times$
&Dedicated survey of graph learning methods, including GCN and GAT, datasets, and graph-based fake review detection challenges
\\

~\citet{maurya2023deceptive}
&2023
&$\checkmark$
&$\triangle$
&$\times$
&$\times$
&$\times$
&Review-, reviewer-, and graph-based features, traditional ML, early DL, datasets, evaluation
\\

~\citet{ennaouri2023machine}
&2023
&$\checkmark$
&$\checkmark$
&$\triangle$
&$\times$
&$\times$
&Systematic review of ML and DL approaches, feature engineering, graph-based methods, benchmark datasets, evaluation metrics, research challenges.
\\

~\citet{duma2024fake}
&2024
&$\checkmark$
&$\checkmark$
&$\checkmark$
&$\triangle$
&$\times$
&Review of traditional methods, deep learning, Transformer/PLM approaches, benchmark datasets, evaluation metrics, challenges, and future research directions.
\\

~\citet{10652819}
&2024
&$\times$
&$\checkmark$
&$\times$
&$\triangle$
&$\times$
&Review of deep learning, benchmark datasets, evaluation metrics, challenges, and future research directions.
\\

~\citet{duma2025analysis}
&2025
&$\times$
&$\checkmark$
&$\times$
&$\times$
&$\times$
&Specialized survey of graph convolutional networks, graph attention networks, heterogeneous GNNs, graph representation learning, graph datasets, evaluation protocols, and open challenges.
\\

~\citet{singh2025deepsurvey}
&2025
&$\triangle$
&$\checkmark$
&$\times$
&$\times$
&$\times$
&Survey of deep learning architectures, Transformer models, early LLM applications, datasets, evaluation metrics, and future research.
\\

~\citet{singh2025deep}
&2025
&$\triangle$
&$\checkmark$
&$\checkmark$
&$\triangle$
&$\times$
&CNN, RNN, attention, Transformer, PLMs, datasets, evaluation, challenges
\\

\textbf{Ours}
&2026
&$\checkmark$
&$\checkmark$
&$\checkmark$
&$\checkmark$
&$\checkmark$
&\textbf{PLM-to-LLM evolution, LLM mechanisms, direct, semantic, and generative detection approaches, augmentation and adversarial learning, future directions}
\\

\bottomrule
\end{tabularx}

\vspace{2mm}

\footnotesize
\textit{Note:} $\checkmark$ denotes broad coverage, $\triangle$ denotes limited discussion, and $\times$ denotes not covered. LLM approaches include prompting, instruction tuning, in-context learning, retrieval-augmented generation, reasoning, parameter-efficient fine-tuning, and agent-based detection.

\end{table*}

Early surveys mainly focused on review spam detection based on handcrafted features and conventional machine learning algorithms.
For example, \citet{dewang2018state} reviewed review spammer detection by summarizing linguistic, behavioral, temporal, metadata, and graph-based features together with supervised, semi-supervised, and unsupervised learning algorithms.
\citet{vidanagama2020deceptive} extended the discussion to deceptive review detection by reviewing linguistic, behavioral, and metadata features, benchmark datasets, and representative traditional learning methods.
As deceptive behaviors became more diverse, early feature-based approaches encountered limitations in capturing high-level semantic patterns and contextual dependencies in user-generated reviews.
This limitation, together with the development of representation learning, motivated the transition from manually designed feature extraction toward deep representation learning frameworks.

With the adoption of deep learning, several surveys broadened their scope to include neural network-based approaches.
\citet{mohawesh2021fake} and \citet{mewada2022research} reviewed feature engineering, traditional machine learning, deep learning models, benchmark datasets, evaluation metrics, and research challenges.
Similarly, \citet{he2022online} reviewed online spam review detection from the perspectives of linguistic, behavioral, and reviewer-related features, while \citet{maurya2023deceptive} summarized review-, reviewer-, and graph-based detection methods together with representative datasets and evaluation protocols.

Recent survey papers have become more specialized.
For example, \citet{duma2025analysis} presented a systematic literature review dedicated to graph neural network-based fake review detection, focusing on graph construction strategies, representative GNN architectures, benchmark datasets, and open challenges.
\citet{singh2025deepsurvey} surveyed deep learning approaches for fake review detection, covering convolutional neural networks, recurrent neural networks, attention mechanisms, Transformer-based models, representative pre-trained language models, and commonly used evaluation benchmarks.
Existing surveys have documented the transition from handcrafted feature engineering to deep learning techniques, and they provide useful references on datasets, evaluation protocols, and representative detection methods.
Table~\ref{tab:survey_comparison} provides a concise comparison of these representative surveys in the literature..

Although existing surveys have advanced the understanding of fake review detection, several research gaps remain.
As shown in Table~\ref{tab:survey_comparison}, previous surveys mainly focus on traditional machine learning, deep learning, graph-based approaches, or early pre-trained language model-based methods.
Some recent surveys discuss Transformer-based models and preliminary LLM applications, but they do not yet provide a fusion-oriented and comprehensive account of LLM-era fake review detection in practical applications.

First, existing surveys rarely connect model evolution with evidence fusion.
Previous studies typically categorize detection approaches by model architecture, while giving less attention to how foundation models reshape the fusion of textual, behavioral, structural, and contextual signals.
The development of contextual representation learning, instruction-following, and generative reasoning requires a new perspective for organizing existing research.

Second, existing reviews lack a unified taxonomy for emerging LLM-based fake review detection methods.
Recent studies have explored direct LLM-based discrimination, LLM-enhanced feature extraction and semantic representation, generative discrimination, and LLM-driven data augmentation and adversarial learning.
These approaches remain fragmented across different research directions and have not been compared in terms of fusion mechanisms, advantages, limitations, and application scenarios.

Third, LLM-generated fake reviews introduce challenges beyond those considered in previous surveys.
The quality and diversity of machine-generated deceptive content raise issues related to source reliability, detection robustness, generalization, trustworthy evaluation, and adversarial adaptation.
These challenges require synthesis from the perspective of LLM-era fake review detection.

To address these gaps, this survey reviews fake review detection from the combined perspective of PLM-to-LLM model evolution and information fusion.
Different from surveys that mainly summarize methods by architecture or traditional detection category, we establish a taxonomy that links evidence sources, fusion levels, and LLM-era detection paradigms.
We review direct discriminative detection, feature extraction and semantic enhancement, generative discriminative detection, and LLM-based data augmentation and adversarial learning.
We further discuss limitations and research directions for reliable fake review detection under AI-generated deceptive content.

\section{Background and Problem Formulation}
\label{sec:background}

\subsection{Survey Methodology}

We conducted a structured literature collection process to identify representative studies on fake review detection.
The goal was not to claim a strict systematic review, but to build a traceable corpus that captures the transition from traditional machine learning to pre-trained language models, large language models, and multi-source information fusion.

\textbf{Literature Search Strategy}.
We searched publications from Web of Science, IEEE Xplore, ACM Digital Library, ScienceDirect, SpringerLink, and Google Scholar.
The search used combinations of the following terms: \textit{fake review, review spam, opinion spam, deceptive review, review spam detection, fake review detection, deceptive review detection, LLM-generated review, multimodal fake review detection, graph-based fake review detection, behavioral fake review detection}.
The search covered studies published from January 1, 2018, to March 27, 2026,
which spans the period from Transformer-based language models to recent 
LLM-based fake review detection.
To improve consistency across databases, we used a common query pattern.
Each query combined a task term, such as \textit{fake review} or \textit{opinion spam}, with a method or evidence source term, such as \textit{BERT}, \textit{large language model}, \textit{graph}, \textit{behavior}, or \textit{multimodal}.
For each candidate paper, we recorded the source database, search term group, publication year, venue type, target task, dataset family, modeling family, evidence sources, reported metrics, and screening decision.
The search log was used to remove duplicate records and to check coding consistency before the final counts were reported.
Because this survey is representative rather than PRISMA-style systematic, the counts below describe the coded corpus rather than all records retrievable from the databases.

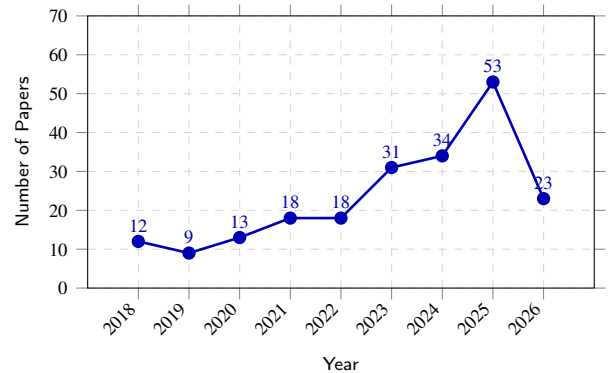
\begin{figure}[pos=htbp]
    \centering
    \begin{tikzpicture}
\begin{axis}[
    width=0.8\columnwidth,
    height=3.6cm,
    scale only axis,
    xmin=2017,
    xmax=2027,
    ymin=0,
    ymax=70,
    xtick={2018,2019,2020,2021,2022,2023,2024,2025,2026},
    xticklabel style={
        rotate=45,
        anchor=east,
        font=\scriptsize
    },
    /pgf/number format/1000 sep={},
    ytick={0,10,20,30,40,50,60,70},
    xlabel={Year},
    ylabel={Number of Papers},
    xlabel style={font=\scriptsize},
    ylabel style={font=\scriptsize},
    tick label style={font=\scriptsize},
    grid=major,
    grid style={dashed,gray!30},
    tick align=outside,
    every axis plot/.append style={
        line width=1pt
    },
    nodes near coords,
    nodes near coords style={
        font=\scriptsize
    },
    nodes near coords align={vertical},
    mark size=2pt
]

\addplot[
    color=blue!70!black,
    mark=*]
coordinates {
(2018,12)
(2019,9)
(2020,13)
(2021,18)
(2022,18)
(2023,31)
(2024,34)
(2025,53)
(2026,23)
};

\end{axis}
\end{tikzpicture}
    \caption{Annual distribution of fake review detection studies included in this survey.}
    \label{fig:year_count_labe}
\end{figure}

\begin{figure}[pos=t]
    \centering
    \begin{tikzpicture}

\begin{axis}[
    width=0.8\columnwidth,
    height=3.6cm,
    scale only axis,
    ybar,
    bar width=18pt,
    ymin=0,
    ymax=120,
    ytick={0,20,40,60,80,100,120},
    ylabel={Number of Papers},
    symbolic x coords={
        ML,
        DL,
        PLM,
        LLM
    },
    xtick=data,
    xticklabel style={
        font=\small,
        align=center
    },
    ymajorgrids=true,
    grid style={dashed,gray!30},
    axis line style={black},
    tick style={black},
    enlarge x limits=0.18,
    nodes near coords,
    every node near coord/.append style={
        font=\small
    }
]

\addplot[
    draw=black,
    fill=blue!55
]
coordinates {
    (ML,92)
    (DL,71)
    (PLM,30)
    (LLM,18)
};

\end{axis}

\end{tikzpicture}
    \caption{Distribution of surveyed papers across different detection paradigms.}
    \label{fig:c_count_labe}
\end{figure}
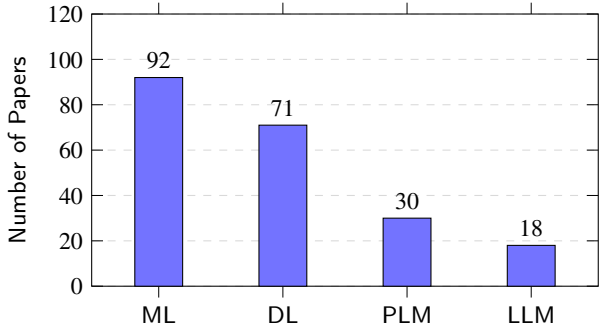

\textbf{Study Selection Criteria}.
The collected publications were screened according to inclusion and exclusion criteria designed to ensure relevance and methodological clarity.
The inclusion criteria were defined as follows:
\begin{enumerate}
    \item Peer-reviewed journal articles and conference papers.
    \item Studies focusing on fake review detection, review spam detection, or deceptive opinion detection.
    \item Studies proposing, developing, or evaluating computational detection methods.
    \item Representative benchmark studies on PLMs, LLMs, multimodal models, graph models, or fusion-based methods that have been applied to fake review detection.
\end{enumerate}
The exclusion criteria were defined as follows:
\begin{enumerate}
    \item Duplicate records.
    \item Non-English publications.
    \item Short papers, editorials, tutorials, patents, and dissertations.
    \item Studies unrelated to fake review detection, such as fake news detection, spam email detection, generic text classification, or image forgery detection.
    \item Papers lacking sufficient methodological descriptions or experimental validation.
\end{enumerate}
Studies from adjacent tasks are excluded from the coded corpus unless they explicitly evaluate fake review detection.
When such studies are cited later, they serve only as methodological analogues for mechanisms such as semantic embedding fusion, graph propagation, adversarial augmentation, or paraphrase resistance.
Screening was conducted in two stages.
Titles and abstracts were first checked for task relevance.
For borderline papers, the method, dataset, and experiment sections were rechecked before inclusion.
No article entered the coded corpus solely because it was cited as background evidence.

Each retained study was coded along two dimensions.
The first dimension records the dominant modeling family, namely machine learning, deep learning, PLM-based modeling, or LLM-based modeling.
The second records the evidence sources and fusion level used by the method, including text-only modeling, text-behavior fusion, text-graph fusion, multimodal fusion, knowledge-enhanced fusion, and decision-level fusion.
This coding scheme supports both the historical analysis of model evolution and the information fusion taxonomy used in this survey.
If a method spans multiple modeling families, the paper is assigned according to the dominant component used for detection, while its evidence sources are recorded separately.
This rule avoids double counting in the paradigm distribution and preserves the fusion information needed for later analysis.

Fusion level is defined by where heterogeneous evidence is combined in the detection pipeline.
Feature-level fusion combines handcrafted linguistic, behavioral, rating, or temporal variables before classification.
Representation-level fusion learns joint embeddings from text, user behavior, graphs, images, or external knowledge.
Graph-level fusion propagates information across reviewer-review-product relations to capture coordinated manipulation.
Decision-level fusion combines predictions or confidence scores from multiple detectors.
Uncertainty-aware fusion further models missing sources, noisy labels, and disagreement among evidence streams.
These definitions are used consistently when methods are compared across ML, DL, PLM, and LLM paradigms throughout this survey.
The literature selection process resulted in three stages.

\textbf{Overview of the Selected Literature}.
The initial search retrieved 359 candidate publications from the selected databases during January 1, 2018 to March 27, 2026.
After removing 71 duplicate records, 288 unique studies remained for screening.
Based on title, abstract, and full-text examination following the predefined criteria, 77 studies were excluded due to irrelevance, insufficient methodological details, or lack of experimental validation.
The final collection contains representative studies spanning traditional machine learning, deep learning, PLM-based methods, LLM-based approaches, and multi-source fusion methods.
The coded corpus contains 211 studies.
By dominant modeling family, 92 studies are ML-based, 71 are DL-based, 30 are PLM-based, and 18 are LLM-based.
Figure~\ref{fig:year_count_labe} illustrates the annual publication trend of the selected literature, showing increased research activity after 2018 and faster growth after the adoption of foundation models.
Figure~\ref{fig:c_count_labe} summarizes the distribution of selected studies across detection paradigms.
Because the 2026 records only cover publications available by early 2026, the 2026 count should be interpreted as a partial-year count rather than a full-year estimate.
The collected literature forms the basis for the taxonomy and comparative analysis presented in the subsequent sections.

\subsection{Task Definition and Scope}

\begin{figure}[pos=htbp]
  \centering
  \begin{overpic}[width=\linewidth]{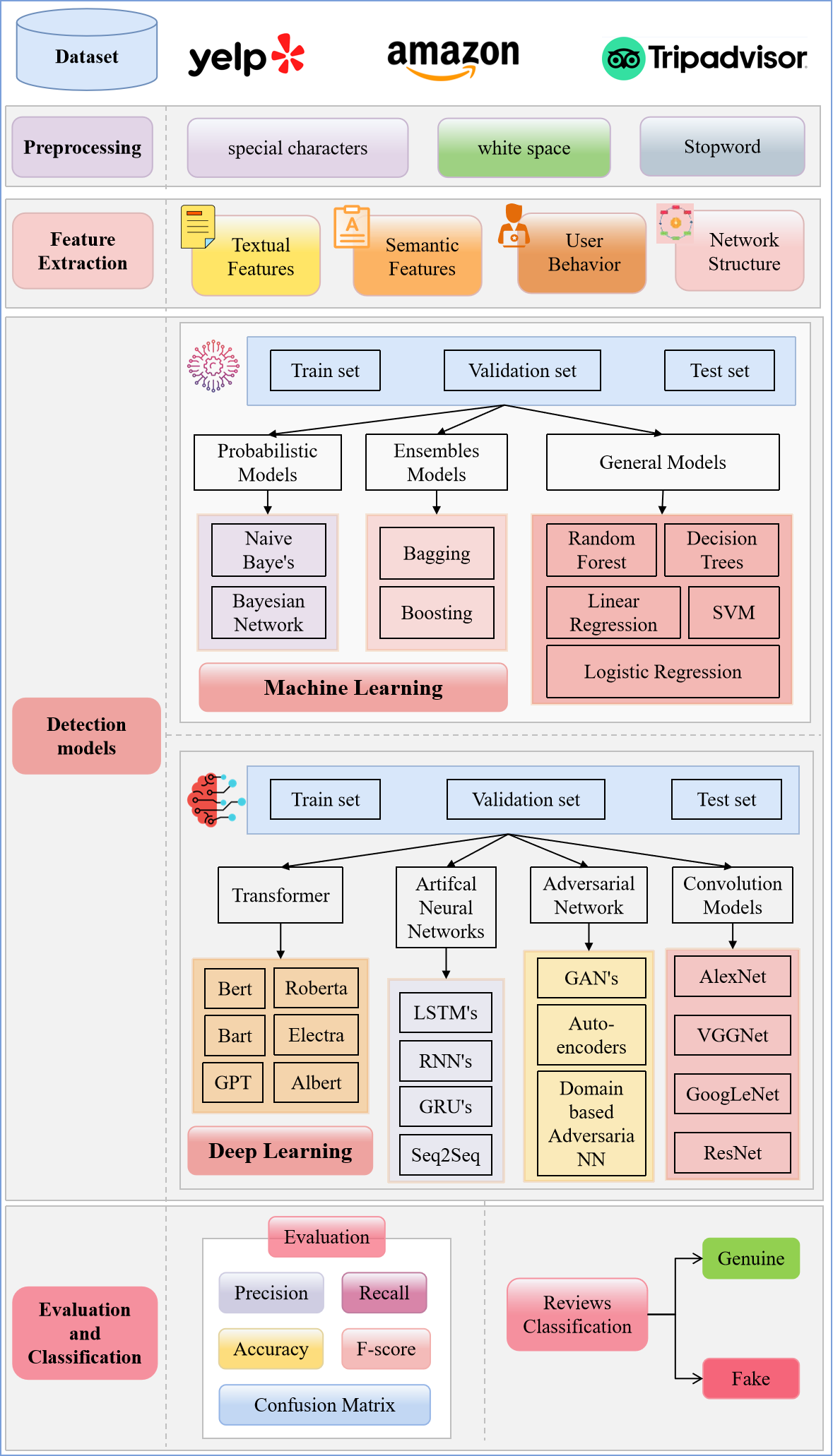}
    \begingroup
    \setlength{\fboxsep}{0pt}
    \endgroup
  \end{overpic}
  \caption{General pipeline of fake review detection, including data preprocessing, feature extraction, detection models, model evaluation and review classification.}
  \label{fig:fake_review_detection_overview}
\end{figure}

Existing studies generally define \textit{fake reviews}, also referred to as \textit{deceptive opinions} or \textit{opinion spam}, as user-generated online content crafted with the intent to deceive, aiming to artificially enhance or damage product reputation, distort perceived product quality, and mislead consumers through deliberate manipulation~\cite{salminen2025decoding}.
Unlike ordinary noisy reviews, fake reviews are strategic and purpose-driven.
Their authors do not write from genuine consumption experiences or personal opinions.
Instead, these reviews are typically produced as part of commercial manipulation or platform-driven influence schemes.

\citet{jindal2008opinion} proposed a widely cited three-category taxonomy of deceptive reviews:

\begin{enumerate}
    \item \textbf{Untruthful Reviews:} Reviews that express sentiment or opinion but are not grounded in genuine consumption experiences, often containing exaggerated or disparaging statements.
    \item \textbf{Brand-Oriented Reviews:} Reviews that target a brand, manufacturer, or seller in the abstract rather than evaluating a specific product or service.
    \item \textbf{Non-Reviews:} Content that includes advertisements, referral links, or information without any substantive evaluative meaning.
\end{enumerate}

This taxonomy clarifies that fake reviews are not merely low-quality or irrelevant content, but fabricated narratives with deliberate deceptive intent.
Their surface characteristics often closely resemble those of genuine reviews, making manual inspection and verification difficult.

Fake review detection research has evolved along two main perspectives, textual and behavioral.
Textual approaches analyze linguistic and stylistic cues to identify deceptive writing patterns.
Behavioral methods exploit anomalous user and merchant activity patterns as indirect signals of manipulation.
Recent research adopts hybrid frameworks that integrate both perspectives and further incorporate graph-based modeling and semantic analysis as complementary tools for generalization and practical deployment.
Figure~\ref{fig:fake_review_detection_overview} illustrates the general pipeline of fake review detection.

\textbf{Textual-level detection} assesses review authenticity by analyzing lexical patterns, syntactic structures, sentiment orientation, and pragmatic cues~\cite{mohawesh2021fake}.
A substantial body of research has identified systematic differences between fake and genuine reviews across multiple highly contextualized linguistic dimensions.
Fake reviews tend to be short, show weak relevance to specific product attributes, exhibit limited sentiment vocabulary, and lack descriptive detail or narrative depth~\cite{wang2022fake,kumar2024exploring}.
They also display a high degree of textual redundancy, with writing styles that are brief and template-like.
At a finer level of linguistic analysis, fake reviews are characterized by excessive lexical repetition, overly embellished sentence structures, frequent use of filler words, and exaggerated or overly generalized lexical choices~\cite{abri2020fake}.
These patterns reflect the cognitive burden of deceptive writing during review generation, which can appear as subtle linguistic and discourse-level anomalies.

Research has also identified consistent psycholinguistic signals in fake reviews.
Fake reviews tend to contain more negations, anxiety-related words, and negative emotion expressions.
They frequently use exaggerated language, abnormal punctuation such as excessive exclamation marks, and place disproportionate emphasis on utilitarian factors such as price~\cite{salminen2025decoding}.
Atypical temporal references and implicit derogation of competitors also serve as identifiable cues.
These psycholinguistic features align well with the contextual modeling capabilities of Transformer-based architectures, providing a basis for building both high-performance and interpretable detection systems~\cite{gupta2024recent}.

Early detection methods relied on manually engineered features such as word frequency counts and sentiment scores, which captured evaluative stance but lacked alignment with specific product attributes~\cite{shehnepoor2021hin}.
Subsequent work shifted toward deep neural networks capable of learning more complex latent semantic patterns from raw text.
These methods perform well on low-complexity fake reviews but lose effectiveness when confronted with highly fluent, human-like writing.
This limitation has become especially pronounced with the emergence of LLM-generated reviews, whose linguistic naturalness now closely approximates that of authentic user-generated content in practice.

\textbf{Behavioral-level detection} focuses on reviewers' historical activity patterns, account interaction modes, temporal behavior sequences, and posting regularities rather than review content itself.
These methods are effective at identifying anomalous users and coordinated manipulation campaigns at scale.
On platforms with anonymous users or heavily obfuscated data, however, behavioral signals tend to be sparse and their interpretability limited.
\citet{martens2019towards} showed that fake reviews are frequently associated with anomalous reviewer behavior, including abnormally high posting frequencies, account activities that deviate from typical consumer usage, and highly skewed rating distributions such as an excessive concentration of five-star or one-star scores.
Graph-based modeling approaches extend behavioral analysis by representing relationships among reviewers, reviews, and merchants as nodes and edges, enabling the discovery of coordinated behavioral patterns and organized manipulation structures~\cite{yu2022graph}.
\citet{sun2024fake} further incorporated both reviewer credibility and merchant credibility into this framework, arguing that fake review generation results from bidirectional interactions between these two parties.
\citet{he2022detecting} showed that products subject to fake review manipulation tend to exhibit higher clustering density in their co-reviewer networks, where the same group of reviewers repeatedly appears across multiple products.
This structural regularity supports the use of relational features for fake review detection.

Fake reviews increasingly exhibit collective and structured patterns at the platform level through coordinated behaviors.
On many platforms, they form reviewer, product, and vendor subgraph structures, with multiple accounts coordinating around the same product.
Reviewers often interact closely, forming tightly-knit clusters, and review bursts frequently coincide with promotional periods or ranking milestones.
These patterns deviate markedly from the distribution of genuine consumer behavior~\cite{salminen2025decoding}.

Hybrid detection approaches integrate textual features with behavioral signals to model the interaction between review content and user activity.
Such methods have demonstrated strong practical utility in platform-level governance and industry applications~\cite{mohawesh2021fake}.
They can simultaneously capture writing anomalies and behavioral irregularities.
These approaches, however, require higher data granularity and labeling costs.
As AI-generated fake reviews become more prevalent, deceptive behaviors increasingly transcend the boundaries of content and behavior, evolving into an adversarial process between generators and detectors.

The emergence of LLMs has changed the nature of fake review generation.
Early template-based fake reviews with extreme sentiment or repetitive phrasing have given way to fluent, contextually coherent narratives that closely mimic authentic user writing~\cite{kreps2022all}.
LLM-generated fake reviews employ strategies such as context alignment, style transfer, and semantic diversification to achieve a high degree of content plausibility~\cite{zellers2019defending}.
As a result, the detection problem has shifted from identifying surface-level textual anomalies to recognizing semantic deception and inferring manipulative intent~\cite{qin2023chatgpt}.
This evolution renders detection methods based on lexical statistics or shallow semantic features increasingly inadequate, and motivates the adoption of more powerful language understanding frameworks.

\subsection{Conventional Fake Review Detection}

Fake review detection methods can be broadly categorized into two developmental stages, the machine learning stage and the deep learning stage.
The former relies on handcrafted features fed into classical classifiers, while the latter employs neural networks to learn semantic representations from raw text and behavioral signals automatically.

\textbf{Machine Learning Stage.}
Before the emergence of deep learning and PLMs, fake review detection mainly relied on traditional machine learning.
Features were manually extracted from review content and reviewer behavior, then fed into supervised classifiers for detection.
Detection performance in this stage depended heavily on the quality of carefully designed feature engineering.
Table~\ref{tab:ml_results_3} summarizes representative machine learning methods for fake review detection.
Early mainstream classifiers included Naive Bayes, Support Vector Machines, and Random Forests.
\citet{lin2014towards} proposed a cross-domain detection model based on sparsity and Bayesian generative models, using part-of-speech tagging and unigram features to capture linguistic and psychological characteristics of fake reviews.
Evaluated on hotel, doctor, and restaurant datasets~\cite{ott2011finding}, the unigram model achieved cross-domain accuracy of 76.1\% in the Turkish review scenario and 77\% in the restaurant domain, but only from 52\% to 64.7\% in the medical domain.
\citet{hernandez2017cross} employed support vector networks combined with latent Dirichlet allocation for feature extraction, evaluating performance across single-domain, mixed-domain, and cross-domain settings.
Their results showed that feature fusion is most effective in single-domain scenarios under controlled settings.

Subsequent work explored feature selection and ensemble strategies to improve the overall model generalization.
\citet{khurshid2018enactment} proposed a two-tier ensemble architecture combining particle swarm optimization with chi-square tests for feature selection, achieving improved classification accuracy on platform and semi-synthetic datasets~\cite{mukherjee2013fake,ott2011finding}.
\citet{sanchez2018character} combined character-level n-gram features with SVM and Naive Bayes classifiers.
\citet{mani2018spam} validated on a gold-standard dataset~\cite{ott2013negative} that stacked ensembles outperform simple voting ensembles, though their performance remained below that of deep learning models.
\citet{nilizadeh2019think} proposed the OneReview model, which identifies review change points from cross-site metadata and integrates them with a Random Forest classifier, achieving notable accuracy improvements at the cost of additional computational time lag.
\citet{li2017bimodal} introduced a coordinated burst detection strategy to capture temporally clustered fake review activity.

A separate line of work addressed the temporal instability of detection models.
\citet{cardoso2018towards} observed that the continual evolution of spam strategies leads to time-varying performance degradation in deployment.
\citet{sanchez2020masking,mohawesh2021analysis} further showed that fake review characteristics shift over time, causing steady performance decline on Yelp datasets~\cite{mukherjee2013fake,rayana2015collective,barbado2019framework}.
These findings highlight the need for models with dynamic adaptability in real-world environments.

\begin{table*}[t]
\centering
\small
\setlength{\tabcolsep}{3pt}
\caption{Performance Summary of Machine Learning-based Fake Review Detection Methods by Dataset Construction Source}
\label{tab:ml_results_3}
\renewcommand{\arraystretch}{1.05}

\begin{tabular}{lllcccc}
\toprule
\textbf{Model} & \textbf{Ref.} & \textbf{Dataset} & \textbf{Acc.} & \textbf{P} & \textbf{R} & \textbf{F1} \\
\midrule

\multicolumn{7}{l}{\textbf{Rule-Based or Platform-Derived Labels}} \\ 
\midrule

DKG-FRD & \cite{Fang2020DynamicKG} & Amazon & 0.9341 & -- & 0.9412 & 0.9304 \\
EBC-MD & \cite{vidanagama2022ontology} & Amazon & 0.8898 & 0.8869 & 0.9090 & 0.8979 \\
MINS & \cite{noekhah2018novel} & Amazon & 0.8900 & -- & -- & -- \\
RF & \cite{he2022detecting} & Amazon & 0.8600 & -- & -- & -- \\

\midrule
\multicolumn{7}{l}{\textbf{Amazon Mechanical Turk Annotation}} \\ 
\midrule

MGSD & \cite{noekhah2020opinion} & OpSpam & 0.9120 & 0.9080 & 0.9000 & 0.9100 \\
Multi-NB & \cite{bhopale2021review} & OpSpam & 0.8813 & 0.8802 & -- & 0.8855 \\
OCRs-RF & \cite{shan2021conflicts} & OpSpam & 0.9290 & 0.9360 & 0.9280 & 0.9320 \\
OneReview & \cite{nilizadeh2019think} & OpSpam & 0.9700 & 0.9100 & 0.9000 & 0.9000 \\
Ramp-One-Class-SVM & \cite{tian2020non} & OpSpam & 0.9213 & -- & -- & -- \\
SE(SVM,NB,RF) & \cite{mani2018spam} & OpSpam & 0.8768 & -- & -- & -- \\
Self-training-NB & \cite{ligthart2021analyzing} & OpSpam & 0.9300 & -- & -- & -- \\
SIPUL & \cite{shunxiang2023building} & OpSpam & 0.8620 & -- & -- & 0.8610 \\
TF-IDF-SGD & \cite{santos2025improving} & OpSpam & 0.9467 & 0.9521 & 0.9329 & 0.9463 \\
VE(SVM,NB,RF) & \cite{mani2018spam} & OpSpam & 0.8743 & -- & -- & -- \\

\midrule
\multicolumn{7}{l}{\textbf{Filtering Algorithm Labels}} \\ 
\midrule

ADWIN & \cite{mohawesh2021analysis} & Yelp-CHI & 0.8675 & -- & -- & -- \\
Co-training (m-f-f) & \cite{wang2020fake} & Yelp-CHI & 0.8445 & 0.8397 & 0.8445 & 0.8189 \\
DATL-FGKD & \cite{tao2026toward} & Yelp-CHI & -- & -- & -- & 0.7300 \\
SIPUL & \cite{shunxiang2023building} & Yelp-CHI & 0.8590 & -- & -- & 0.8520 \\
ADWIN & \cite{mohawesh2021analysis} & Yelp-NYC & 0.9214 & -- & -- & -- \\
DATL-FGKD & \cite{tao2026toward} & Yelp-NYC & -- & -- & -- & 0.6900 \\
ADWIN & \cite{mohawesh2021analysis} & Yelp-ZIP & 0.7352 & -- & -- & -- \\
DATL-FGKD & \cite{tao2026toward} & Yelp-ZIP & -- & -- & -- & 0.7100 \\
SIPUL & \cite{shunxiang2023building} & Yelp-ZIP & 0.8670 & -- & -- & 0.8660 \\

\midrule
\multicolumn{7}{l}{\textbf{Human Annotation or Curated Benchmarks}} \\ 
\midrule

GADM & \cite{ruan2020gadm} & Yelp & 0.8330 & -- & -- & 0.8620 \\
OneReview & \cite{nilizadeh2019think} & Yelp & 0.9700 & 0.9100 & 0.9000 & 0.9000 \\
PSRaT-AdaBoost & \cite{barbado2019framework} & Yelp & -- & -- & -- & 0.8200 \\
Ramp-One-Class-SVM & \cite{tian2020non} & Yelp & 0.7437 & -- & -- & -- \\
RF & \cite{shan2021conflicts} & Yelp & 0.9290 & 0.9360 & 0.9280 & 0.9320 \\
Self-training-NB & \cite{ligthart2021analyzing} & Yelp & 0.7300 & -- & -- & -- \\
XGboost & \cite{sihombing2019fake} & Yelp & -- & 0.9900 & 0.9900 & 0.9900 \\

\bottomrule
\end{tabular}
\end{table*}

\begin{table*}[t]
\centering
\small
\setlength{\tabcolsep}{3pt}
\caption{Performance Comparison of Deep Learning Methods by Dataset Construction Source, Part 1}
\label{tab:dl_results_1}
\renewcommand{\arraystretch}{1.05}

\begin{tabular}{lllccccc}
\toprule
\textbf{Model} & \textbf{Ref.} & \textbf{Dataset} & \textbf{Acc.} & \textbf{P} & \textbf{R} & \textbf{F1} & \textbf{AUC} \\
\midrule
\multicolumn{8}{l}{\textbf{Rule-Based or Platform-Derived Labels}} \\ \midrule
Bi-LSTM & \cite{baishya2021safer} & Amazon & 0.9630 & -- & -- & -- & -- \\
CNN & \cite{hajek2020fake} & Amazon & 0.8130 & -- & -- & 0.8060 & 0.8790 \\
CNN & \cite{sikchi2024fake} & Amazon & 0.9500 & -- & -- & 0.9500 & -- \\
DFFNN & \cite{hajek2020fake} & Amazon & 0.8280 & -- & -- & 0.8250 & 0.8930 \\
DHMFRD-TER & \cite{duma2024dhmfrd} & Amazon & 0.9880 & 0.9880 & 0.9860 & 0.9870 & -- \\
DNN & \cite{hajek2023fake} & Amazon & 0.8289 & -- & -- & 0.8290 & 0.8970 \\
GEmFakeABSA & \cite{chauhan2026gemfakeabsa} & Amazon & 0.9240 & 0.8370 & 0.8535 & 0.8451 & -- \\
GNN-HHO & \cite{Oak2024MetaheuristicGNN} & Amazon & 0.8760 & 0.8681 & 0.8867 & 0.8774 & -- \\
HIN-RNN & \cite{shehnepoor2021hin} & Amazon & -- & 0.8500 & 0.9000 & 0.8700 & -- \\
LRGCL & \cite{yao2024fake} & Amazon & -- & -- & 0.9055 & 0.8595 & 0.9746 \\
MTL & \cite{melleng2023multi} & Amazon & -- & -- & -- & 0.7310 & -- \\
fakeRoBERTa & \cite{SALMINEN2022102771} & Amazon & -- & 0.9700 & 0.9700 & 0.9700 & -- \\
\midrule
\multicolumn{8}{l}{\textbf{Amazon Mechanical Turk Annotation}} \\ \midrule
BiLSTMWF-POS-I & \cite{liu2020incorporating} & OpSpam & 0.8520 & 0.8450 & 0.9090 & 0.8760 & -- \\
FakeGAN & \cite{aghakhani2018detecting} & OpSpam & 0.8910 & 0.9800 & 0.8100 & -- & -- \\
OPCNN & \cite{zhao2018towards} & OpSpam & 0.8450 & -- & 0.8103 & 0.8284 & -- \\
PV-DAE & \cite{fahfouh2020pv} & OpSpam & 0.9250 & 0.9125 & 0.9358 & 0.9240 & -- \\
RoBERTa-TF-IDF-ALSTM & \cite{jakhar2026explainable} & OpSpam & 0.9906 & 0.9876 & 0.9938 & 0.9907 & -- \\
ScoreGAN & \cite{shehnepoor2021scoreganfraudreviewdetector} & OpSpam & 0.7726 & 0.7160 & -- & -- & 0.8767 \\
\midrule
\multicolumn{8}{l}{\textbf{Filtering Algorithm Labels}} \\ \midrule
DHM & \cite{duma2023deep} & Yelp-CHI & 0.9950 & 0.9770 & 0.9610 & 0.9650 & -- \\
DHMFRD-TER & \cite{duma2024dhmfrd} & Yelp-CHI & 0.9870 & 0.9560 & 0.9340 & 0.9240 & -- \\
HFAN & \cite{yuan2019learning} & Yelp-CHI & -- & 0.4887 & -- & -- & 0.8324 \\
IFML(MVAE,LLMs)-BAF & \cite{Xu2025Interpretable} & Yelp-CHI & -- & -- & -- & -- & 0.9443 \\
LRGCL & \cite{yao2024fake} & Yelp-CHI & -- & -- & 0.8415 & 0.7720 & 0.9291 \\
EMVDLM & \cite{mohawesh2023explainable} & Yelp-NYC & -- & 0.9630 & 0.8416 & 0.9009 & -- \\
Ensemble & \cite{mohawesh2023explainable} & Yelp-NYC & 0.9693 & -- & -- & -- & -- \\
HFAN & \cite{yuan2019learning} & Yelp-NYC & -- & 0.5382 & -- & -- & 0.8478 \\
IFML(MVAE,LLMs)-BAF & \cite{Xu2025Interpretable} & Yelp-NYC & -- & -- & -- & -- & 0.9378 \\
LRGCL & \cite{yao2024fake} & Yelp-NYC & -- & -- & 0.7774 & 0.5590 & 0.7869 \\
BeGPX & \cite{manaskasemsak2023fake} & Yelp-NYC & -- & 0.9700 & -- & -- & -- \\
CNN & \cite{jain2019deceptive} & Yelp-Zip & 0.6640 & 0.6700 & 0.6500 & -- & -- \\
EMVDLM & \cite{mohawesh2023explainable} & Yelp-ZIP & -- & 0.8318 & 0.7786 & 0.8043 & -- \\
Ensemble & \cite{mohawesh2023explainable} & Yelp-ZIP & 0.8318 & -- & -- & -- & -- \\
HFAN & \cite{yuan2019learning} & Yelp-Zip & -- & 0.6235 & -- & -- & 0.8728 \\
IFML(MVAE,LLMs)-BAF & \cite{Xu2025Interpretable} & Yelp-Zip & -- & -- & -- & -- & 0.9072 \\
BeGPX & \cite{manaskasemsak2023fake} & Yelp-Zip & -- & 0.9700 & -- & -- & -- \\
\midrule
\bottomrule
\end{tabular}
\end{table*}

\begin{table*}[t]
\centering
\small
\setlength{\tabcolsep}{3pt}
\caption{Performance Comparison of Deep Learning Methods by Dataset Construction Source, Part 2}
\label{tab:dl_results_2}
\renewcommand{\arraystretch}{1.05}

\begin{tabular}{lllccccc}
\toprule
\textbf{Model} & \textbf{Ref.} & \textbf{Dataset} & \textbf{Acc.} & \textbf{P} & \textbf{R} & \textbf{F1} & \textbf{AUC} \\
\midrule
\multicolumn{8}{l}{\textbf{Human Annotation or Curated Benchmarks}} \\ \midrule
A-LSTM & \cite{xu2024hybridizeda} & Yelp & 0.9090 & 0.8870 & 0.8790 & -- & -- \\
HIN-RNN & \cite{shehnepoor2021hin} & Yelp & -- & 0.8100 & 0.8200 & 0.8100 & -- \\
Meta-MGDN & \cite{2024-40717} & Yelp & -- & -- & -- & -- & 0.8150 \\
ScoreGAN & \cite{shehnepoor2021scoreganfraudreviewdetector} & Yelp & 0.8476 & 0.6516 & -- & -- & 0.8878 \\
CNN & \cite{hajek2020fake} & Yelp-Doctor & 0.8835 & -- & -- & 0.9100 & 0.9460 \\
DFFNN & \cite{hajek2020fake} & Yelp-Doctor & 0.8621 & -- & -- & 0.8930 & 0.9320 \\
AEDA & \cite{you2018attribute} & Yelp-Hotel & 0.8000 & 0.8390 & 0.7420 & 0.7870 & -- \\
CNN & \cite{hajek2020fake} & Yelp-Hotel & 0.8725 & -- & -- & 0.8720 & 0.9450 \\
DFFNN & \cite{hajek2020fake} & Yelp-Hotel & 0.8956 & -- & -- & 0.8960 & 0.9510 \\
EUPHORIA & \cite{andresini2022review} & Yelp-Hotel & -- & -- & -- & 0.5920 & 0.8130 \\
OpCNN & \cite{zhao2018accuratedeceptiveopinionspam} & Yelp-Hotel & 0.8450 & -- & 0.8103 & 0.8284 & -- \\
RFPPO-EM & \cite{yao2021ensemble} & Yelp-Hotel & -- & -- & -- & 0.7225 & -- \\
AEDA & \cite{you2018attribute} & Yelp-Restaurant & 0.7560 & 0.8240 & 0.6510 & 0.7280 & -- \\
CNN & \cite{hajek2020fake} & Yelp-Restaurant & 0.8980 & -- & -- & 0.9010 & 0.9650 \\
DFFNN & \cite{hajek2020fake} & Yelp-Restaurant & 0.8831 & -- & -- & 0.8840 & 0.9530 \\
EUPHORIA & \cite{andresini2022review} & Yelp-Restaurant & -- & -- & -- & 0.3720 & 0.7080 \\
\bottomrule
\end{tabular}
\end{table*}

Beyond supervised methods, researchers also explored unsupervised and semi-supervised approaches to address the scarcity of reliable labeled data in practice.
\citet{akram2018finding} proposed an unsupervised model that identifies duplicate reviews using cosine similarity, classifying reviews below a similarity threshold as genuine without human intervention.
\citet{dong2018unsupervised} introduced a joint topic and sentiment model built on an improved LDA, feeding extracted topic and sentiment features into Random Forest and SVM classifiers.
Experiments on a Yelp dataset showed improvements over POS-based, character n-gram, and unigram baselines.
\citet{noekhah2020opinion} proposed a graph-structured unsupervised model integrating explicit and implicit features, with experiments on crowdsourced and AMT datasets~\cite{ott2011finding} confirming that structured feature fusion improves detection performance.
\citet{deng2017semi} applied K-Means clustering to generate pseudo-labels from semantic similarity scores, achieving 88.1\% average accuracy on JD.com electronic product reviews, though performance degraded on short texts of fewer than 20 words.
\citet{hai2016deceptive} employed Laplacian Regularized Logistic Regression to learn from unlabeled samples, with experiments on the Yelp dataset demonstrating improvements in both precision and recall under limited labeled data settings~\cite{tian2020non,wang2020fake,ligthart2021analyzing}.

Traditional machine learning methods share several common limitations.
They rely heavily on manually engineered features, which constrain generalization and reduce robustness against evolving deceptive patterns.
Most approaches focus on surface-level statistical differences in writing style while neglecting the semantic and intentional dimensions of deception.
Reviews are typically treated as independent samples, without modeling discourse-level consistency or long-term identity-mimicking behaviors.
These limitations indicate that fake review detection cannot rely solely on observable surface features, but must engage with deeper semantic reasoning.

\textbf{Deep Learning Stage.}
Compared with traditional machine learning methods, deep neural networks can learn higher-order semantic representations and latent patterns directly from raw text.
Table~\ref{tab:dl_results_1} and Table~\ref{tab:dl_results_2} summarize representative deep learning methods for fake review detection.
Convolutional Neural Networks have been widely applied to extract local features and sentence-level representations from review text~\cite{li2015learning}.
\citet{zhao2018towards} adopted a word-order-preserving CNN with word embeddings and specific pooling strategies for short text classification, achieving high efficiency and accuracy.
\citet{fahfouh2020pv} combined CNN with denoising autoencoders to build a hybrid model capable of extracting embedded features from reviews.

Recurrent Neural Networks capture sequential information in text and have been applied to document-level fake review detection.
\citet{wang2018detecting} proposed an LSTM-based system incorporating dictionary features, using a three-layer MLP combined with LSTM for enhanced feature representation.
\citet{liu2020incorporating} combined bidirectional LSTM with part-of-speech tagging, first-person pronoun features, and GloVe embeddings to produce document-level review representations.
\citet{jain2019deceptive} introduced a hierarchical CNN-GRN architecture that uses recurrent networks to handle long sequences.
\citet{zhang2018dri} proposed the DRIRCNN model, which extracts contextual features through convolutional layers and captures sequence dependencies through recurrent layers, demonstrating the effectiveness of hybrid architectures on AMT and deceptive review datasets.

\begin{table*}[t]
\centering
\caption{Representative Public Datasets for Fake Review Detection}
\label{tab:dataset_summary}
\renewcommand{\arraystretch}{1.15}
\setlength{\tabcolsep}{4pt}

\begin{tabularx}{\textwidth}{
p{2.2cm}
p{0.9cm}
p{2.2cm}
p{2.2cm}
p{3.6cm}
X
}
\toprule

\textbf{Dataset} &
\textbf{Ref.} &
\textbf{Total Reviews} &
\textbf{Domain} &
\textbf{Available Information} &
\textbf{Typical Usage} \\

\midrule

Amazon &
\cite{jindal2008opinion} &
5.4M+ &
E-commerce &
Text, rating, product information, reviewer ID, timestamp &
Traditional ML, feature engineering, behavioral analysis
\\

OpSpam &
\cite{ott2011finding,ott2013negative} &
1,600 (positive) 
1,600 (negative) &
Hotel &
Review text &
Text-based fake review detection, PLM evaluation, cross-domain studies
\\

Yelp-CHI &
\cite{mukherjee2013fake} &
67,395 &
Restaurant (Chicago) &
Text, rating, reviewer profile, business information &
Behavior-aware detection, feature engineering
\\

Yelp-NYC &
\cite{rayana2015collective} &
359,052 &
Restaurant 
(New York City) &
Text, rating, reviewer, business, social interactions &
Graph learning, reviewer relation modeling, GNN-based detection
\\

Yelp-ZIP &
\cite{rayana2015collective} &
608,598 &
Restaurant 
(ZIP-code regions) &
Text, rating, reviewer, business, graph structure &
Heterogeneous graph learning, graph neural networks
\\

Dianping &
\cite{Li7023420} &
$\sim$2.1M &
Restaurant &
Text, rating, reviewer, timestamp &
Weakly supervised learning, PU learning, large-scale evaluation
\\

Yelp-Hotel &
\cite{zhang2018dri} &
2,160 &
Hotel &
Review text &
Cross-domain deceptive review detection
\\

Yelp-Restaurant &
\cite{zhang2018dri} &
401 &
Restaurant &
Review text &
Cross-domain deceptive review detection
\\

Yelp-Doctor &
\cite{zhang2018dri} &
556 &
Healthcare &
Review text &
Cross-domain deceptive review detection
\\

\bottomrule

\end{tabularx}
\vspace{2mm}
\footnotesize
\textit{Note:} Available Information denotes the information provided by each dataset, including review text, ratings, reviewer profiles, business metadata, timestamps, and graph structures. Typical Usage summarizes the most common research scenarios in the literature.
\end{table*}

Attention mechanisms enable adaptive weighting of heterogeneous feature representations and improve the model's ability to focus on informative signals.
\citet{wang2017detecting} employed an attention network to dynamically weight linguistic and behavioral features, using an MLP for behavioral pattern extraction and a CNN for textual representation, achieving effective multimodal fusion.
\citet{yuan2019learning} proposed a hierarchical fusion attention network that learns representations at both product and user levels, integrating orthogonal decomposition techniques with TransH embeddings to model complex relationships.

Researchers have incorporated external information such as structured knowledge, behavioral network relations, and user or merchant credibility into fake review detection.
This line of work integrates textual semantics with structural constraints to detect more complex manipulation behaviors.
Unlike approaches that focus mainly on linguistic semantics or language-model generation mechanisms, these methods use knowledge graphs, social graph structures, and behavioral sequences to obtain richer contextual representations.

One of the most representative knowledge-enhancement strategies involves the construction of knowledge graphs, in which entities such as reviews, products, and users, together with their semantic relations, are explicitly modeled to incorporate higher-order structural information.
\citet{Fang2020DynamicKG} proposed a dynamic knowledge graph-based fake review detection method that first extracts multiple types of entities including review texts, products, and users from textual and temporal information, constructs a dynamic knowledge graph with temporal dimensions, and defines diverse relational indicators to strengthen the identification of deceptive reviews.
Experimental evaluations demonstrate that this model outperforms traditional text classification approaches by explicitly encoding the evolving relational context surrounding each review.
\citet{duma2025analysis} further pointed out through systematic survey that graph-structured learning improves detection performance by modeling the relationships among reviews, users, and products, and highlighted the potential of graph-based models to capture behavioral correlations in multi-source heterogeneous structural data.

Structural constraints derived from social networks and behavioral graphs have also been extensively studied as complementary sources of discriminative evidence.
\citet{Cheng2024FakeReviewerGNN} proposed a social-context graph-based framework for fake reviewer detection, in which relationships between users and reviews as well as relationships among users themselves are modeled as multiple subgraphs, and graph neural networks are employed to learn behavioral and relational patterns.
This approach identifies fake review publishers and deceptive patterns, achieving high detection accuracy on discussion forum datasets.
Methods based on Graph Convolutional Networks, which learn mutual information and adjacency relationships between users and reviews, have similarly been shown to capture review behavior patterns and structural anomalies that are difficult to detect through text analysis alone.
\citet{yu2022graph} summarize the importance and challenges of such approaches in modeling structural constraints over reviews and associated metadata, providing a reference for graph-based fake review detection research.

In addition to structural relationships, behavioral credibility has been introduced as a form of knowledge enhancement signal that bridges textual and non-textual sources of evidence.
\citet{sun2024fake} proposed jointly modeling review textual content with sequential features derived from user and merchant behaviors, using behavioral credibility scores and activity sequence patterns to reflect potential fraudulent activities.
By integrating textual features with behavioral knowledge within a unified detection framework, this strategy yields consistent performance improvements on datasets such as Dianping across multiple benchmark datasets and demonstrates the value of fusing heterogeneous information sources for fake review detection.

Knowledge-enhanced and structure-constrained methods depart from text-centric detection approaches by explicitly modeling the relational context in which fake reviews are produced and disseminated.
The integration of knowledge graphs, graph neural networks, and behavioral modeling provides complementary discriminative signals that strengthen detection against group-level and organized deceptive behaviors in practical settings.

Generative Adversarial Networks provide a framework for both generating and detecting fake reviews.
\citet{aghakhani2018detecting} proposed the FakeGAN model, which uses semi-supervised learning to generate samples consistent with the distribution of real data, addressing the problem of data scarcity.
\citet{you2018attribute} applied GANs to integrate product, reviewer, and review attribute information, using a domain classifier to handle the cold-start problem.
\citet{tang2020generating} used GANs to generate synthetic behavioral features for new users, enabling cold-start fake review detection.

Graph Neural Networks exploit structural relationships among reviewers, products, and social connections.
\citet{guo2021deep} employed GNNs to integrate embeddings of both incidental and stable relationships, effectively identifying spam reviewers.
Multi-feature fusion models combine linguistic, behavioral, and structural signals for detection.
\citet{cao2020deceptive} integrated fine-grained and coarse-grained features extracted through a two-layer neural network and LDA, performing final classification with an SVM.
\citet{noekhah2018novel} proposed an unsupervised multi-iteration network architecture that simultaneously detects individual reviews and collective behavior patterns.

\begin{table*}[t]
\centering
\caption{Comparison of Label Construction Methods in Public Fake Review Datasets}
\label{tab:dataset_construction}
\renewcommand{\arraystretch}{1.18}
\setlength{\tabcolsep}{4pt}

\begin{tabularx}{\textwidth}{
p{2.5cm}
p{2.2cm}
X
X
X
}
\toprule

\textbf{Construction Method} &
\textbf{Dataset} &
\textbf{Label Construction} &
\textbf{Main Limitation} &
\textbf{Potential Impact}
\\

\midrule
\multicolumn{5}{l}{\textbf{Rule-based Techniques}}\\
\midrule

Rule-based &
Amazon~\cite{jindal2008opinion}
&
Duplicate and near-duplicate reviews are treated as spam proxies, supplemented by manually identified Type-2 and Type-3 spam.
&
Duplicate reviews represent only a subset of deceptive reviews and cannot capture sophisticated opinion spam.
&
Models may overfit duplicate patterns rather than genuine deceptive behavior, limiting generalization to real-world scenarios.
\\

\midrule
\multicolumn{5}{l}{\textbf{Amazon Mechanical Turk (AMT)}}\\
\midrule

AMT &
OpSpam~\cite{ott2011finding}
&
Truthful hotel reviews are collected from TripAdvisor, while deceptive reviews are generated by Amazon Mechanical Turk workers under controlled instructions.
&
Artificially generated deceptive reviews differ from naturally occurring spam and contain only positive reviews.
&
Models may learn AMT writing styles instead of authentic deception cues, reducing ecological validity.
\\

AMT &
Negative OpSpam~\cite{ott2013negative}
&
Truthful negative reviews are collected from online travel websites, while deceptive negative reviews are written by AMT workers.
&
Still relies on simulated deception despite including negative opinions.
&
Supports sentiment-balanced evaluation but remains different from real-world spam campaigns.
\\

\midrule
\multicolumn{5}{l}{\textbf{Filtering Algorithms}}\\
\midrule

Filtering &
Yelp-CHI~\cite{mukherjee2013fake}
&
Labels are derived from Yelp's proprietary filtering system (filtered vs.\ recommended reviews).
&
The filtering algorithm is a black box and filtered reviews are only approximate ground truth.
&
Provides realistic large-scale data but introduces label uncertainty and platform dependency.
\\

Filtering &
Yelp-NYC / Yelp-ZIP~\cite{rayana2015collective}
&
Platform-filtered reviews are regarded as deceptive, while recommended reviews are treated as truthful.
&
Ground-truth labels completely rely on Yelp's filtering mechanism and may overlook hidden spam.
&
Widely adopted benchmark for graph learning, but conclusions remain tied to Yelp-specific assumptions.
\\

Filtering &
Dianping~\cite{Li7023420}
&
Reviews filtered by Dianping are treated as positive samples, while the remaining reviews are considered unlabeled for positive-unlabeled learning.
&
Platform filtering provides incomplete supervision and hidden spam remains unlabeled.
&
Supports realistic large-scale weakly supervised learning but introduces uncertain labels.
\\

\midrule
\multicolumn{5}{l}{\textbf{Human Annotation}}\\
\midrule

Human Annotation &
Opinion Dataset~\cite{li2011learning}
&
Reviews are manually annotated according to review content, reviewer profiles, and historical behaviors.
&
Manual annotation is expensive, subjective, and difficult to scale.
&
Provides high-quality labels but limited dataset size and domain coverage.
\\

Human Annotation &
Yelp-Hotel / Yelp-Restaurant / Yelp-Doctor~\cite{zhang2018dri}
&
Reviews are manually verified and annotated by domain experts based on predefined deceptive review criteria.
&
Small-scale datasets with limited domain diversity.
&
Suitable for controlled cross-domain evaluation but insufficient for training large-scale neural models.
\\

\bottomrule

\end{tabularx}

\vspace{2mm}

\footnotesize
\textit{Note:} Public fake review datasets are generally constructed using four strategies: rule-based heuristics, Amazon Mechanical Turk (AMT), platform filtering algorithms, and human annotation. Each construction strategy introduces different levels of label reliability and potential bias, which directly influence the evaluation and generalization ability of fake review detection models.

\end{table*}

Despite their advances over traditional methods, CNN and LSTM-based models still face fundamental limitations.
They perform adequately on low-complexity fake reviews with extreme sentiment or templated language, but struggle with semantically deceptive reviews that closely resemble authentic writing.
Content-only approaches ignore anomalous correlations among review targets and fail to detect coordinated manipulation campaigns where the same group of reviewers targets multiple products.
Semantic understanding in these models remains at the syntactic level, lacking reasoning capacity and world knowledge.
As a result, such models can assess whether a review appears genuine but cannot evaluate deeper aspects such as the legitimacy of user intent or the consistency between a review and the reviewer's identity~\cite{ren2017neural}.
These limitations motivated the adoption of PLMs, shifting fake review detection from shallow feature matching to deeper semantic understanding.

\subsection{Datasets and Evaluation Metrics}

Datasets used in fake review detection research are generally constructed through four approaches, filtering-algorithm-based methods, crowdsourcing platforms, manual annotation, and rule-based methods.
Table~\ref{tab:dataset_summary} presents representative public datasets used in fake review detection, including their scale, domains, available information, and typical research applications.
Table~\ref{tab:dataset_construction} compares label construction strategies, showing how rule-based heuristics, platform filtering, crowdsourcing, and manual annotation introduce different assumptions and annotation limitations.
Table~\ref{tab:dataset_characteristics} summarizes the commonly reported characteristics and challenges of representative benchmarks from four perspectives, label construction, dataset bias, generalization issues, and benchmark considerations.
These analyses show that dataset construction directly affects label reliability, model robustness, cross-domain generalization, and the fairness of experimental comparisons.

\begin{table*}[!htbp]
\centering
\caption{Characteristics and Challenges of Representative Public Fake Review Datasets}
\label{tab:dataset_characteristics}
\renewcommand{\arraystretch}{1.18}
\setlength{\tabcolsep}{4pt}

\begin{tabularx}{\textwidth}{
p{2.3cm}
X
X
X
X
}

\toprule

\textbf{Dataset} &
\textbf{Label Construction Characteristics} &
\textbf{Known Dataset Bias} &
\textbf{Reported Generalization Issues} &
\textbf{Benchmark Considerations}
\\

\midrule

Amazon &
Spam labels are primarily inferred from duplicate and near-duplicate reviews, with limited manually verified spam cases.
&
Biased toward duplicate-review patterns and Amazon-specific review behaviors.
&
Cross-platform generalization remains largely unexplored.
&
Large-scale benchmark, but heuristic labels cannot fully represent genuine deceptive reviews.
\\

OpSpam &
Truthful reviews are collected from TripAdvisor, whereas deceptive reviews are generated by Amazon Mechanical Turk workers under controlled settings.
&
Limited to hotel reviews with experimentally generated deception.
&
Models trained on AMT-generated reviews may not transfer well to naturally occurring opinion spam.
&
Widely adopted gold-standard benchmark for text-based deception detection despite its synthetic nature.
\\

Yelp-CHI &
Labels are obtained from Yelp's proprietary filtering system by distinguishing filtered and recommended reviews.
&
Restaurant-domain dataset relying on Yelp-specific filtering mechanisms.
&
Cross-domain and cross-platform robustness remains limited.
&
Provides realistic user behaviors, although ground-truth labels depend on an undisclosed filtering algorithm.
\\

Yelp-NYC / Yelp-ZIP &
Platform-filtered reviews are regarded as deceptive while recommended reviews are treated as truthful.
&
Platform-specific reviewer interactions and graph structures dominate the data distribution.
&
Models may rely on Yelp-specific relational patterns instead of domain-independent deception cues.
&
Large-scale benchmark widely used for graph-based fake review detection.
\\

Dianping &
Platform-filtered reviews are treated as positive samples, while remaining reviews are considered unlabeled for positive-unlabeled learning.
&
Chinese restaurant reviews with language- and platform-specific behavioral characteristics.
&
Cross-language and cross-platform transferability has not been extensively validated.
&
Representative benchmark for weakly supervised and PU-learning settings rather than fully supervised learning.
\\

Opinion Dataset &
Labels are manually assigned based on review content, reviewer profiles, and historical behaviors.
&
Small-scale dataset with limited domain diversity.
&
Limited evidence for large-scale or cross-domain evaluation.
&
Provides relatively reliable manual annotations but is unsuitable for training large neural models.
\\

Yelp-Hotel / Yelp-Restaurant / Yelp-Doctor &
Reviews are manually verified according to predefined deceptive-review criteria.
&
Small datasets covering only several application domains.
&
Useful for controlled cross-domain experiments but insufficient for evaluating foundation models.
&
Frequently adopted for domain adaptation studies rather than large-scale benchmark evaluation.
\\

\bottomrule

\end{tabularx}

\vspace{2mm}
\footnotesize
\textit{Note:} Rather than providing subjective quality ratings, this table summarizes representative characteristics and commonly acknowledged challenges of public fake review datasets reported in the original studies, including label construction strategies, dataset-specific biases, generalization concerns, and benchmark considerations.

\end{table*}

\textbf{Filtering-algorithm-based datasets.}
The Yelp CHI dataset was collected by \citet{mukherjee2013fake} between 2004 and 2012, containing 67,365 restaurant and hotel reviews from Chicago.
Reviews were labeled as fake or genuine by Yelp's proprietary spam filtering system.
User behavioral features were collected by analyzing advertising data and internal website logs, including geographic information, IP addresses, session logs, and network connections.
\citet{rayana2015collective} adopted the same methodology and collected two datasets from Yelp.com covering 2004 to 2015, namely \textit{Yelp-NYC} and \textit{Yelp-ZIP}.
Yelp-NYC contains 359,052 reviews and Yelp-ZIP includes 608,598 reviews, with an average review length of 130.6 words.
\citet{Li7023420} constructed a Chinese dataset using a review filtering algorithm, containing 9,765 reviews with an average length of 85.5 words.
These datasets have become widely adopted benchmarks for evaluating fake review detection methods because they provide large-scale annotations collected from online platforms.
A common limitation of filtering-algorithm-based datasets is that the labeling mechanisms rely on undisclosed proprietary algorithms whose exact criteria are not publicly available.

\textbf{Crowdsourcing-based datasets.}
Amazon Mechanical Turk enables large-scale data collection by distributing tasks to anonymous online workers.
Although humans often struggle to accurately distinguish fake reviews from genuine ones, they can be instructed to deliberately generate deceptive reviews, which are then used as labeled samples.
\citet{ott2011finding} constructed a dataset of 800 hotel reviews from Chicago, collecting 400 genuine reviews from TripAdvisor and generating 400 fake reviews via crowdsourcing.
\citet{ott2013negative} later extended this to a dataset of 1,600 reviews, half of which were labeled as fake.
\citet{Li2014Towards} adopted the same approach to create a dataset of 3,032 reviews.
A shared limitation of crowdsourcing-based datasets is that their distribution differs from that of platform reviews, which constrains their ecological validity.

\textbf{Manually annotated datasets.}
\citet{li2011learning} constructed a dataset based on manual annotation guided by 30 handcrafted rules.
Three undergraduate student volunteers independently labeled each review as fake or genuine, and a majority voting strategy was applied.
Reviews labeled as fake by at least two annotators were considered deceptive, yielding a dataset of 6,000 reviews, of which 1,398 were labeled as fake.
\citet{Ren2014PositiveUnlabeled} constructed a dataset of 3,000 reviews containing 712 fake reviews using a similar procedure.
Manual annotation is labor-intensive and highly time-consuming.
Human annotators generally exhibit limited accuracy in identifying fake reviews~\cite{ott2013negative}, and manually labeled datasets therefore tend to contain annotation noise to some extent.

\textbf{Rule-based datasets.}
\citet{jindal2008opinion} proposed an approach based on Amazon reviews, identifying three types of duplicated reviews likely to be fake: different reviewer IDs posting reviews for the same product, the same reviewer ID posting reviews for different products, and different reviewer IDs posting reviews for different products.
Textual similarity was computed using Jaccard distance, and reviews with a similarity score above 0.9 were labeled as fake.
The resulting dataset contains 5.8 million reviews, approximately 55,000 of which were labeled as deceptive.
\citet{Fornaciari2014FakeAmazon} and \citet{Hammad2015ArabicSpam} followed similar rule-based principles and constructed datasets containing 6,819 and 20,848 book and hotel reviews, respectively.
\citet{barbado2019framework} crawled review data from Yelp.com and labeled reviews based on content and user behavioral features, resulting in 9,653 fake reviews and 20,828 genuine reviews.

Rule-based methods eliminate the need for manual annotation and can generate large-scale labeled datasets efficiently.
They inevitably introduce noise, however.
In the dataset of \citet{jindal2008opinion}, reviews are labeled as fake when multiple users review the same product or the same user submits multiple reviews for the same product.
Such assumptions are not always reliable, as a consumer may legitimately post multiple reviews for the same product due to network disruptions or platform management issues.
These reviews may nonetheless be incorrectly labeled as deceptive.

The evaluation of fake review detection systems is inherently challenging, as it must simultaneously account for class imbalance, heterogeneous feature sources, model uncertainty, and decision interpretability.
Consequently, recent studies have moved beyond conventional single-metric evaluation toward a broader multi-dimensional assessment framework that captures complementary aspects of model performance in practice.
In the literature, evaluation metrics can be broadly categorized into three groups, classification performance metrics, consistency and reliability metrics, and interpretability metrics.

\textbf{Classification performance metrics} for fake review detection models primarily rely on metrics derived from the confusion matrix.
Although Accuracy provides an overall measure of correctness, it is insufficient in imbalanced settings, where genuine reviews typically dominate the dataset.
To address this limitation, prior studies therefore consistently emphasize class-sensitive metrics.

Precision reflects the reliability of positive predictions, namely reviews predicted as fake, which is necessary for reducing false accusations against genuine users.
Recall, also known as the True Positive Rate, measures the proportion of actual fake reviews correctly identified, directly affecting the effectiveness of fraud detection systems.
F1-score provides a harmonic balance between Precision and Recall and is therefore widely adopted as a primary evaluation metric in benchmark datasets such as Amazon, Yelp, and OpSpam.

Threshold-independent metrics are also used.
AUC-ROC evaluates class separability across varying decision thresholds, making it suitable for comparing models with different calibration characteristics.
True Negative Rate complements recall by measuring the model's ability to correctly identify genuine reviews, thereby providing a more balanced assessment across classes.
To further address the limitations of F1-score under severe class imbalance, recent studies advocate the use of Matthews Correlation Coefficient, which incorporates all elements of the confusion matrix and provides a correlation-based evaluation.

\textbf{Consistency and reliability metrics}.
With the rise of LLM-based and prompt-driven detection frameworks, evaluation has expanded to include model stability and reliability, as such models may produce stochastic outputs even under identical inputs.
To capture these properties, recent studies introduce several complementary evaluation metrics.
Consistency across repeated trials evaluates whether the model produces stable predictions under repeated inference.
Mean confidence and its standard deviation capture both the average prediction confidence and its variability, thereby reflecting the model’s calibration and uncertainty.
Cohen’s Kappa Score measures the agreement between predicted labels and ground truth while adjusting for chance agreement, providing a more reliable estimate of prediction consistency than raw accuracy.

\textbf{Interpretability metrics}.
As fake review detection systems are deployed in high-stakes applications, such as e-commerce moderation and reputation management, model interpretability has become a practical requirement.
Among existing approaches, SHAP is widely used to quantify the contribution of linguistic cues, behavioral signals, and other features to the final prediction.
It supports both local explanations for individual instances and comprehensive global feature importance analysis.

For multi-source fusion methods, evaluation should also test whether a detector remains reliable when one evidence source is missing, noisy, or adversarially corrupted.
Source-wise ablation, calibration metrics such as expected calibration error, and robustness tests under missing metadata or sparse user histories can reveal whether performance is driven by genuine complementary evidence or by a single dominant source.
In summary, the evaluation of fake review detection systems has evolved toward a framework that more comprehensively integrates performance, reliability, interpretability, and source-level robustness.

\section{PLM-Based Fake Review Detection Methods}
\label{sec:PLM}

PLMs have improved text understanding in fake review detection.
Compared with approaches based on handcrafted feature engineering or shallow neural networks, PLMs learn contextualized semantic representations from large-scale unlabeled corpora, enabling them to capture latent characteristics of deceptive reviews in terms of wording patterns, emotional expressions, and discourse structures~\cite{10.1145/3503162.3503169,liu2019roberta}.
Depending on the functional role of PLMs within the detection framework, existing methods can be grouped into four categories, text semantic encoders, end-to-end fine-tuned classifiers, feature fusion and enhancement modules, and interpretable detection components.
Existing PLM-based studies differ mainly in model architecture, semantic modeling mechanism, and the role assigned to pretrained representations.
Table~\ref{tab:plm_results} summarizes PLM-based and hybrid fake review detection methods.

\begin{table*}[t]
\centering
\small
\setlength{\tabcolsep}{3pt}
\caption{Performance Comparison of Representative PLM-based Methods by Dataset Construction Source}
\label{tab:plm_results}
\renewcommand{\arraystretch}{1.05}

\begin{tabular}{lllcccc}
\toprule
\textbf{Model} &
\textbf{Ref.} &
\textbf{Dataset} &
\textbf{Acc.} &
\textbf{P} &
\textbf{R} &
\textbf{F1} \\
\midrule

\multicolumn{7}{l}{\textbf{Rule-Based or Platform-Derived Labels}}\\
\midrule

MBO-DeBERTa &
\cite{geetha2025high} &
Amazon &
0.7800 &
0.7700 &
0.7800 &
0.7700 \\

DeBERTa-v3 &
\cite{NARAYAN2025100285} &
Amazon &
0.9500 &
-- &
-- &
0.9800 \\

\midrule
\multicolumn{7}{l}{\textbf{Amazon Mechanical Turk Annotation}}\\
\midrule

BERT &
\cite{10.1145/3497701.3497714} &
OpSpam &
0.9100 &
-- &
-- &
0.9300 \\

MBO-DeBERTa &
\cite{geetha2025high} &
OpSpam &
0.9100 &
0.9100 &
0.9000 &
0.9000 \\

RoBERTa-LSTM &
\cite{MOHAWESH2024250} &
OpSpam &
0.9603 &
0.9148 &
0.9931 &
0.9536 \\

\midrule
\multicolumn{7}{l}{\textbf{Human Annotation or Curated Benchmarks}}\\
\midrule

RoBERTa &
\cite{10.1145/3503162.3503169} &
Yelp &
0.6900 &
-- &
-- &
0.6900 \\

BERT &
\cite{10.1145/3497701.3497714} &
Yelp &
0.7300 &
-- &
-- &
0.7300 \\

XLNet &
\cite{shajalal2024mattersexplanationsexplainablefake} &
Yelp &
0.9349 &
0.9278 &
0.9654 &
0.8044 \\

BSTC &
\cite{lu2023bstc} &
Yelp-Hotel &
0.9344 &
0.9064 &
0.9688 &
0.9336 \\

RoBERTa &
\cite{Hadi2025RoBERTaFakeReview} &
Yelp-Hotel &
0.8625 &
0.8800 &
0.8700 &
0.8600 \\

DeBERTa-v3 &
\cite{NARAYAN2025100285} &
Yelp-Hotel &
0.8100 &
-- &
-- &
0.8200 \\

BSTC &
\cite{lu2023bstc} &
Yelp-Restaurant &
0.9125 &
0.8837 &
0.9500 &
0.9157 \\

DeBERTa-v3 &
\cite{NARAYAN2025100285} &
Yelp-Restaurant &
0.9400 &
-- &
-- &
0.9400 \\

BSTC &
\cite{lu2023bstc} &
Yelp-Doctor &
0.9286 &
0.9706 &
0.9167 &
0.9429 \\

\bottomrule
\end{tabular}

\end{table*}

\subsection{PLMs and Semantic Modeling Mechanisms}

The fundamental architectures of PLMs are predominantly based on the Transformer, whose key advantage lies in modeling dependencies among all tokens in a sequence simultaneously through the self-attention mechanism~\cite{vaswani2017attention}.
In fake review detection, this mechanism enhances the model's ability to capture long-range dependencies.
Fake reviews often conceal deceptive intent through cross-sentence narrative techniques, for instance by maintaining surface-level emotional consistency across multiple sentences while embedding factual inconsistencies.
The Transformer's global attention enables effective capture of inter-sentential semantic relationships and logical contradictions.
Multi-head attention and hierarchical stacking further allow the model to assign differentiated weights to key semantic nodes within reviews, such as concrete experience descriptions, sentiment-bearing expressions, or referential terms, helping distinguish genuine experiential statements from exaggerated or fabricated language patterns.

Contextual embedding is a defining feature that distinguishes PLMs from traditional static word embeddings.
Unlike Word2Vec~\cite{mikolov2013efficient} and GloVe~\cite{pennington2014glove}, which map each word to a fixed vector, PLMs generate highly adaptive and context-aware dynamic representations that adjust according to the surrounding context~\cite{devlin2019bert,Peters2018ELMo}.
This property is particularly valuable in fake review detection, as it enables models to capture semantic shifts and contextual inconsistencies that often characterize deceptive reviews.
PLMs also support multi-granularity feature modeling at the word, sentence, and paragraph levels, which helps the model comprehend the overall semantic logic of a review.
Dynamic contextual adaptation further improves transferability across domains, allowing PLMs to remain sensitive to semantic nuances in reviews from different platforms or product categories.

The application of PLMs in fake review detection has evolved from early shallow semantic modeling based on RNNs and CNNs to deep contextual modeling driven by Transformer architectures.
With the emergence of BERT, RoBERTa, and ELECTRA, PLMs have become the foundational infrastructure for this task~\cite{devlin2019bert,clark2020electra}.

\paragraph{BERT and its variants.}
BERT~\cite{devlin2019bert}, proposed by the Google AI Language team, is a bidirectional Transformer-based PLM that for the first time enables context-aware bidirectional encoding.
Its Masked Language Model objective learns contextual information from both directions, while the Next Sentence Prediction task captures inter-sentence relationships.
In fake review detection, BERT enables deep semantic modeling of review texts, capturing subtle linguistic differences in syntax, sentiment, and semantics that are indicative of deceptive writing.

RoBERTa optimizes BERT's pre-training process by removing the NSP task, introducing dynamic masking, and training on larger corpora, achieving stronger and more reliable generalization performance in fake review detection~\cite{liu2019roberta}.
It demonstrates greater resilience to cross-domain reviews and diverse user expression styles, capturing subtle semantic deception patterns across complex deployment environments.
ALBERT reduces model size through parameter sharing and factorized embedding parameterization, maintaining BERT's representational capacity while significantly decreasing complexity, making it suitable for resource-constrained deployment scenarios.
ELECTRA introduces a Replaced Token Detection mechanism that trains the model to distinguish real tokens from plausible replacements generated by a small generator network~\cite{clark2020electra}.
This pre-training objective aligns closely with the intrinsic logic of fake review detection, as it simulates the subtle emotional or factual substitutions often found in deceptive reviews, making ELECTRA particularly effective at detecting AI-generated or carefully crafted fake content.
DeBERTa further enhances semantic dependency modeling through a Disentangled Attention mechanism and Enhanced Position Embedding strategy, assigning attention separately to the semantic and positional encodings of words~\cite{he2020deberta}.
This design captures sentence structural information more accurately and helps more effectively identify inter-sentence inconsistencies or templated expressions common in deceptive reviews, especially in long-text and multi-turn review scenarios that require extended context.

\paragraph{Knowledge-enhanced models.}
ERNIE~\cite{sun2019ernie} integrates entities, semantic relations, and world knowledge into language representations through a knowledge-enhanced pre-training mechanism.
In the context of fake review detection, this allows the model to identify factual errors and inconsistencies within reviews by grounding textual representations in structured external knowledge.
ERNIE also strengthens inter-sentence relationship learning, facilitating the detection of hidden logical contradictions in deceptive reviews.

\paragraph{Adaptation strategies for fake review detection.}
Directly applying general-purpose PLMs to fake review detection often fails to fully capture complex domain-specific characteristics of review corpora.
Researchers have proposed several adaptation strategies to address this gap.
Multi-task learning jointly trains fake review detection alongside related tasks such as sentiment classification and topic identification, enabling the model to capture richer semantic and emotional representations~\cite{ruder2017overview,melleng2023multi}.
Task-specific fine-tuning on labeled fake review datasets further focuses the model on recognizing semantic disguises, logical contradictions, and template-like expressions~\cite{shawon2023bengali}.
Domain-Adaptive Pre-training fine-tunes a general PLM on domain-specific corpora such as e-commerce or hotel reviews, improving alignment with domain-specific linguistic patterns and vocabulary distributions~\cite{gururangan2020don,catelli2023new}.
This strategy performs well in cross-domain transfer scenarios, though it relies on large-scale domain corpora and high computational resources~\cite{dong2024once}.
Data augmentation and adversarial training enhance model sensitivity to diverse linguistic patterns by generating synthetic fake review samples or applying adversarial perturbations~\cite{liu2025data}.
Augmentation expands training data and improves cross-domain generalization, while adversarial training strengthens robustness against highly disguised reviews under challenging real-world conditions at the cost of increased computational overhead.

\subsection{PLMs as Semantic Encoders}

Before PLMs were widely adopted for fake review detection, the field primarily relied on manually engineered linguistic features and shallow learning models, whose performance was constrained by limited representational capacity and poor domain generalization~\cite{ott2011finding}.
With the advent of models such as BERT, researchers began to employ PLMs as powerful contextual semantic representation encoders, using contextual representations as input to conventional classifiers or shallow neural networks for authenticity prediction.
In this paradigm, PLM parameters are kept frozen and the \texttt{[CLS]} embedding or sentence-level hidden states serve as semantic representations of reviews.

\citet{10.1145/3503162.3503169} systematically evaluated BERT, RoBERTa, ALBERT, and DistilBERT as semantic encoders on the OpSpam and Yelp datasets.
On the Yelp dataset, BERT-based representations achieved an F1 score of approximately 0.88, outperforming traditional CNN models at around 0.82 and LSTM models at around 0.84.
In Chinese-language scenarios, fake review detection faces additional challenges including word segmentation ambiguity, colloquial expressions, and sparse domain-specific vocabulary.
\citet{app12073338} employed BERT-wwm to extract semantic representations from Chinese e-commerce reviews and fed them into a multilayer perceptron classifier, achieving Accuracy of 0.89 and F1 of 0.87, outperforming TF-IDF and sentiment feature baselines.
\citet{Cui_2021} further showed that whole-word masking BERT exhibits stable generalization across multiple Chinese review datasets.
On the Amazon product review dataset, \citet{Hadi2025RoBERTaFakeReview} utilized RoBERTa to extract contextual semantic vectors combined with a simple classifier, achieving Accuracy and F1 scores exceeding 0.90, an improvement of approximately 5 to 8 percentage points over word-embedding-based models.

Considering computational constraints in practical applications, some studies treat PLM outputs as static semantic features combined with traditional classifiers or lightweight neural networks.
\citet{lu2023bstc} extracted BERT embeddings on the Yelp dataset and concatenated them with TextCNN features, achieving an F1 score of approximately 0.89 while significantly reducing training costs.
Similar approaches have been applied to multilingual and low-resource fake review detection tasks, improving deployability with limited performance loss~\cite{MOHAWESH2024250}.
Cross-platform and cross-category evaluations indicate that PLM-based semantic encoding approaches suffer from relatively modest performance degradation under cross-domain evaluation, with F1 scores typically remaining above 0.80~\cite{gupta2024recent}.

Using PLMs as semantic encoders represents a transition from handcrafted features toward deep semantic modeling, elevating F1 scores to the range from 0.85 to 0.90 on mainstream datasets such as OpSpam, Yelp, and Amazon.
This setting does not fully exploit the task adaptation capabilities of pre-trained models, however, and its ability to identify subtle and context-dependent deceptive reviews remains limited.
This limitation has directly motivated the development of end-to-end fine-tuning approaches.

\subsection{PLMs as Fine-tuned Classifiers}

Building upon static semantic encoding, subsequent studies have increasingly adopted end-to-end fine-tuning, in which PLMs are directly employed as the core discriminative models for fake review detection~\cite{duma2023deep}.
Unlike encoder-based approaches with frozen parameters, this paradigm updates PLM parameters through backpropagation from the downstream task, enabling better adaptation to the specific characteristics of fake review corpora.
Review texts are fed directly into models such as BERT or RoBERTa, and the \texttt{[CLS]} representation is passed through a fully connected layer or softmax classifier to predict review authenticity~\cite{devlin2019bert}.
This eliminates the need for manually designed classifier architectures, allowing semantic modeling and discriminative learning to be jointly optimized.

BERT is among the earliest PLMs widely adopted for end-to-end fine-tuning in fake review detection.
On the OpSpam dataset, fine-tuned BERT models achieve Accuracy above 0.90 and F1 values ranging from approximately 0.88 to 0.91, significantly outperforming traditional CNN and LSTM-based models~\cite{sun2020finetuneberttextclassification,ott2011finding}.
On the Dianping review dataset, several studies report F1 scores from 0.89 to 0.92 after fine-tuning BERT, with stable performance on long-form reviews and implicitly deceptive content~\cite{electronics13214322}.
RoBERTa further advances end-to-end fine-tuning performance across text classification tasks.
Fine-tuning RoBERTa on the Yelp fake review dataset yields Accuracy and F1 scores exceeding 0.93, an improvement of approximately 2 to 3 percentage points over fine-tuned BERT, demonstrating stronger task-specific adaptation in complex detection scenarios~\cite{liu2019roberta}.

In non-English and multilingual scenarios, end-to-end fine-tuning has also shown strong effectiveness.
\citet{app12073338} fine-tuned BERT-zh and RoBERTa-zh on a Chinese e-commerce review dataset, achieving F1 scores from 0.88 to 0.91 and outperforming non-fine-tuned semantic encoding approaches.
In low-resource language settings, fine-tuning multilingual BERT or regional language models such as Indic-BERT has yielded F1 scores above 0.85, validating the broad applicability of end-to-end fine-tuning in cross-lingual fake review detection~\cite{De_2025}.

End-to-end fine-tuned PLMs consistently achieve F1 scores above 0.90 on mainstream fake review datasets, making this setting one of the most effective approaches within the PLM-based detection framework.
Two limitations remain, however.
Model performance is highly sensitive to the scale and quality of labeled data, leading to notable degradation in low-resource or cross-domain scenarios.
Fine-tuned models also typically function as black-box systems, offering limited interpretability of their decision-making processes, which poses challenges for platform governance and risk management applications.

\subsection{PLMs as Feature Fusion Modules}

Textual semantic information alone is insufficient for many fake review detection settings.
Fake reviews may contain linguistic anomalies, abnormal reviewer behavior, irregular temporal patterns, inconsistent ratings, and imbalanced sentiment or topical distributions.
Recent studies therefore treat PLMs as feature fusion and enhancement modules.
In these frameworks, PLM representations are combined with behavioral, temporal, sentiment, graph, or visual features to improve generalization under complex deployment conditions by effectively leveraging complementary information from multiple heterogeneous evidence sources.

\paragraph{Fusion of textual semantics and sentiment features.}
Deceptive reviews often exhibit extreme sentiment distributions and overly consistent emotional expressions~\cite{ott2011finding}.
\citet{lu2023bstc} proposed the BSTC model, which employs BERT to extract semantic embeddings from Yelp reviews while incorporating sentiment features derived from sentiment lexicons and emotion classifiers.
The heterogeneous features are fused through convolutional neural networks for final classification.
The model achieves an F1 score of approximately 0.89, outperforming a BERT-only fine-tuning baseline by about two percentage points.
On the Amazon product review dataset, similar sentiment and semantic fusion approaches reduce misclassification of sentiment-camouflaged fake reviews, yielding Accuracy and F1 scores beyond 0.90 under diverse conditions~\cite{Hadi2025RoBERTaFakeReview}.

\paragraph{Fusion of textual semantics and reviewer behavior features.}
Reviewer behavior features such as posting frequency, rating distributions, and account activity levels are important signals for identifying fake reviews~\cite{10.1145/2487575.2487580}.
\citet{sun2024fake} constructed a hybrid framework on the Yelp dataset where BERT encodes review text semantics and the resulting representations are concatenated with behavioral features including historical rating patterns and inter-review time intervals, achieving an F1 score of 0.92, an improvement of approximately 3 to 4 percentage points over text-only models.
Experiments on the OpSpam dataset further indicate that semantic and behavioral fusion models are effective in identifying reviews with natural-sounding text but anomalous behaviors, especially under class-imbalanced settings and practical scenarios.~\cite{10.1145/3503162.3503169}.

\paragraph{Multimodal and structural feature integration.}
Some studies extend PLM-based frameworks to multimodal fake review detection by jointly modeling textual, visual, and metadata features.
In datasets containing product images, textual semantic features from BERT are combined with visual features from ResNet and fused for authenticity prediction, consistently outperforming unimodal text-based approaches with F1 scores above 0.90~\cite{veluru2025multimodal}.
For organized and group-based fraudulent behaviors, structured group detection algorithms such as SGDCTH construct buyer and seller heterogeneous networks and use collaborative training over node attributes and adjacency relations to identify coordinated spamming groups, reducing false positives~\cite{Zhang2023SpammerGroup}.
Heterogeneous graph attention networks fuse review text, user attributes, and relational weights.
Partition-based graph embedding methods introduce structural priors through contrastive learning.
These results indicate that structural knowledge constraints can be more effective than text-only features under data sparsity or high-noise conditions in complex real-world scenarios~\cite{gupta2024recent,Oak2024MetaheuristicGNN}.

Feature fusion approaches improve detection robustness by incorporating complementary signals beyond text, but they also increase model complexity and impose higher demands on data collection, annotation, and computational resources.
Constructing knowledge graphs and structured relational data is often costly, and in domains where entity relationships are sparse, the effectiveness of graph-based constraints may deteriorate.

\subsection{Limitations of PLM-Based Methods}

Although PLMs have improved fake review detection, several limitations constrain practical deployment.

\paragraph{Vulnerability to AI-generated fake reviews.}
With the adoption of generative AI such as the GPT series, fake reviews have become more fluent and coherent.
PLMs are primarily trained for contextual representation and word prediction, making it difficult to capture subtle anomalies in highly realistic AI-generated reviews that mimic genuine user writing style and complex behavioral interaction patterns~\cite{zellers2019defending}.
A fine-tuned BERT or RoBERTa model may correctly classify reviews that exhibit obvious templated phrasing or emotional exaggeration, but fail on reviews generated by LLMs that are semantically coherent and stylistically indistinguishable from authentic writing.
This limitation is structural because PLMs learn statistical patterns from training data, while AI-generated fake reviews can deliberately avoid those patterns and render learned discriminative features obsolete.

\paragraph{Cross-domain generalization.}
PLMs are often fine-tuned on domain-specific corpora such as restaurant reviews, and their performance degrades when transferred to different domains or platforms such as hotels or electronics.
Domain-specific vocabulary, stylistic differences, and shifts in contextual logic prevent the model from effectively detecting deceptive signals outside the training distribution.
Empirical results show that F1 scores can drop by 5 to 15 percentage points under cross-domain evaluation even for strong fine-tuned models~\cite{gupta2024recent}, highlighting the need for stronger domain adaptation strategies under realistic.

\paragraph{Long-text and contextual modeling constraints.}
Basic PLMs are prone to context truncation or semantic confusion when handling long documents or complex sentence structures due to fixed-length input constraints.
Advanced models such as DeBERTa partially alleviate these issues, but remain constrained by the quadratic computational complexity of self-attention mechanisms.
This limits the applicability of PLM-based methods to platforms where reviews are lengthy or structured as multi-turn dialogues.

\paragraph{Reliance on labeled data.}
End-to-end fine-tuning approaches are highly sensitive to the scale and quality of labeled data, which directly affects learning stability and generalization.
In low-resource or newly emerging review domains where labeled fake reviews are scarce, model performance degrades substantially due to limited domain-specific supervision signals.
Semi-supervised and domain-adaptive approaches partially address this issue but introduce additional training complexity and require careful hyperparameter tuning in practice.

\paragraph{Limited interpretability.}
Fine-tuned PLMs typically function as black-box systems, providing predictions without explicit reasoning about why a review is classified as fake.
This lack of interpretability poses challenges for platform governance, regulatory compliance, and user trust, where explainable detection decisions are required in deployment environments.
Attention visualization and probing methods offer partial insights but do not constitute rigorous explanations of model behavior.

These limitations indicate that PLM-based methods are insufficient as standalone solutions for fake review detection in the current setting of AI-generated content.
They require integration with reasoning capabilities, external knowledge, and adaptive learning mechanisms, motivating the large language model-based detection methods discussed in the following section.

\section{Large Language Models in Fake Review Detection}
\label{sec:LLM}

\subsection{Evolution and Core Mechanisms of LLMs}

Generative artificial intelligence has made LLMs central to recent fake review detection research.
Compared with conventional PLMs, LLMs offer larger-scale semantic modeling, stronger task generalization, and instruction-following ability with enhanced adaptability.
Table~\ref{tab:llm_results} summarizes representative LLM era fake review detection results, including LLM-based methods, fusion components, and reported baseline rows from the same studies.

\begin{table*}[t]
\centering
\small
\setlength{\tabcolsep}{3pt}
\caption{Performance Comparison of Representative LLM Era Fake Review Detection Methods and Reported Baselines by Dataset Construction Source}
\label{tab:llm_results}
\renewcommand{\arraystretch}{1.05}

\begin{tabular}{lllccccc}
\toprule
\textbf{Model} &
\textbf{Ref.} &
\textbf{Dataset} &
\textbf{Acc.} &
\textbf{P} &
\textbf{R} &
\textbf{F1} &
\textbf{AUC} \\
\midrule

\multicolumn{8}{l}{\textbf{Rule-Based or Platform-Derived Labels}}\\
\midrule

BiLSTM\_Word2Vec &
\cite{computers13100264} &
Amazon &
0.9846 &
0.9873 &
0.9794 &
0.9833 &
-- \\

BiLSTM &
\cite{Nawara2025DualPhaseLLM} &
Amazon &
0.7730 &
0.7680 &
0.7700 &
0.7580 &
-- \\

Hybrid &
\cite{s25227048} &
Amazon &
-- &
0.8640 &
0.7560 &
0.8030 &
0.9180 \\

KE-MLLM &
\cite{chen2026ke} &
Amazon &
0.9320 &
0.8950 &
0.9080 &
0.9280 &
-- \\

XSMF-LLMIC &
\cite{Wang2026LLMIC} &
Amazon &
0.9375 &
-- &
-- &
0.9378 &
0.9412 \\

FraudSquad &
\cite{liu2025detectingllmgeneratedspamreviews} &
Amazon, Llama 3 &
-- &
0.9099 &
0.9781 &
-- &
0.9994 \\

FraudSquad &
\cite{liu2025detectingllmgeneratedspamreviews} &
Amazon, Qwen2 &
-- &
0.9236 &
0.9902 &
-- &
0.9998 \\

FraudSquad &
\cite{liu2025detectingllmgeneratedspamreviews} &
Amazon, Qwen DSR1 &
-- &
0.8945 &
0.9573 &
-- &
0.9993 \\

AGFRDCP &
\cite{luo2026ai} &
Amazon, GPT 2 &
0.9092 &
0.9210 &
0.8951 &
0.9079 &
-- \\

AGFRDCP &
\cite{luo2026ai} &
Amazon, Qwen DSR1 &
0.8387 &
0.8642 &
0.8037 &
0.8329 &
-- \\

\midrule
\multicolumn{8}{l}{\textbf{Amazon Mechanical Turk Annotation}}\\
\midrule

BERTbase &
\cite{Liyanage2024AIEnhanced} &
OpSpam &
-- &
-- &
-- &
0.9783 &
-- \\

LLaMA 3.1-Finetuning &
\cite{miah2025hiddenplainsightevaluation} &
OpSpam &
0.9225 &
-- &
-- &
0.9224 &
-- \\

LLaMA 3.1-Inference &
\cite{miah2025hiddenplainsightevaluation} &
OpSpam &
0.6293 &
-- &
-- &
0.6247 &
-- \\

\midrule
\multicolumn{8}{l}{\textbf{Filtering Algorithm Labels}}\\
\midrule

KE-MLLM &
\cite{chen2026ke} &
Yelp-CHI &
0.9470 &
0.9080 &
0.9210 &
0.9430 &
-- \\

\midrule
\multicolumn{8}{l}{\textbf{Human Annotation or Curated Benchmarks}}\\
\midrule

FraudSquad &
\cite{liu2025detectingllmgeneratedspamreviews} &
Yelp &
-- &
0.3367 &
0.3800 &
-- &
0.7032 \\

AGFRDCP &
\cite{luo2026ai} &
Yelp, Llama 2 &
0.9834 &
0.9828 &
0.9840 &
0.9834 &
-- \\

LLMChaos &
\cite{FAN2025101521} &
Yelp-Hotel &
0.9527 &
0.9455 &
-- &
0.9478 &
0.9530 \\

XSMF-LLMIC &
\cite{Wang2026LLMIC} &
Yelp-Hotel &
0.9588 &
-- &
-- &
0.9589 &
0.9553 \\

XAI-LDistilBERT &
\cite{bdcc9080205} &
Yelp-Restaurant &
0.9725 &
0.9725 &
0.9756 &
0.9724 &
-- \\

\bottomrule
\end{tabular}

\end{table*}

The GPT series, proposed by OpenAI, employs autoregressive language modeling to generate fluent natural language text~\cite{brown2020language,radford2019language}.
GPT established a foundation for LLMs by supporting multi-task generalization and contextual understanding.
Building on this foundation, instruction tuning and reinforcement learning from human feedback enable the model to better comprehend task intentions, produce controllable responses, and exhibit structured reasoning ability~\cite{achiam2023gpt}.
In contrast to the GPT family, Meta's LLaMA adopts efficient training strategies and parameter optimization to deliver strong language understanding within a comparatively lightweight architecture, making it suitable for academic research and downstream task adaptation~\cite{touvron2023llama}.
The GLM series takes a different approach by adopting a unified framework that combines bidirectional encoding with autoregressive decoding, thereby achieving strong multi-task generalization and multimodal potential~\cite{du2021glm}.

These models rely on deep Transformer architectures to capture long-range dependencies and contextual semantic relationships.
Pretrained on massive corpora spanning web text, books, and question-answering datasets, LLMs acquire extensive world knowledge, language patterns, and logical reasoning rules.
Through explicit task descriptions or carefully designed prompts, LLMs can be directed toward specific tasks such as fake review detection without requiring additional parameter updates, enabling effective zero-shot and few-shot predictions.
Compared with traditional machine learning and deep neural network approaches, LLMs model semantic consistency, discourse logic, and intent correlations at a deeper level, providing support for detecting fluent and evolving deceptive reviews~\cite{papageorgiou2024survey}.

The application of LLMs in fake review detection, however, exhibits an inherent duality.
Their generative capabilities allow them to produce highly realistic fabricated reviews that closely mimic authentic user writing, thereby intensifying the challenge of detection.
At the same time, their semantic reasoning capabilities provide powerful tools for identifying subtle linguistic anomalies and inferring manipulative intent.
This tension between generation and detection defines the central dynamic of LLM research in this domain, and motivates the diverse methodological approaches reviewed in the following subsections.

\subsection{LLM-based Direct Discriminative Detection Methods}

In the most straightforward application setting, LLMs are treated as general-purpose LLM-as-Judge systems that directly output authenticity labels through natural language prompts or instruction-based reasoning.
Such approaches are typically implemented under zero-shot or few-shot settings, avoiding explicit model training and offering flexibility and transferability.
Unlike traditional supervised learning methods, prompt-driven approaches do not rely on parameter updates, but instead guide the model to perform classification or reasoning tasks through carefully designed natural language instructions.
As a result, they demonstrate notable advantages in low-resource, cross-domain, and rapid-deployment scenarios.

\citet{bader2023detecting} examined ChatGPT-generated fake reviews and reported that their fluency and sentiment expression closely resemble human-written reviews, while deviations remain in personalization, emotional fluctuation, and repetition.
\citet{zhao2025ai} further showed through large-scale empirical analysis that AI-generated fake reviews exhibit stable but subtle statistical patterns in lexical diversity, emotional intensity, and syntactic structures.
Together with evidence that fake product reviews generated by LLMs are difficult for humans and machines to distinguish~\cite{meng2025large}, these findings make prompt-driven fake review detection an increasingly consequential research topic.

The most direct prompt-driven strategy invokes LLMs in zero-shot or few-shot settings to classify reviews as genuine or fake.
\citet{lu2023large} observed that with carefully designed prompts or a small number of exemplars, LLMs can generate reviews that are stylistically, emotionally, and logically closer to real reviews, while effectively bypassing existing detection models.
This form of adversarial generation enabled by prompt engineering transforms fake reviews from random, noise-like fabrications into strategically targeted and goal-oriented outputs.
\citet{NARAYAN2025100285} evaluated the zero-shot detection capabilities of models such as ChatGPT-3.5, GPT-4, and LLaMA2 on Amazon and hotel review datasets using instruction templates of the form ``determine whether this review is fake''.
Their results showed that under unsupervised conditions, LLMs typically achieved accuracy rates below 45\%, substantially lagging behind fine-tuned DeBERTa-v3 models that exceeded 96\% accuracy.
This performance gap highlights the limitations of direct prompt-based reasoning when confronting sophisticated fake reviews in complex environments.

Similar conclusions were drawn from the MAiDE-up dataset proposed by \citet{ignat2024maideupmultilingualdeceptiondetection}, which contains 10,000 multilingual fake hotel reviews generated by GPT-4.
Experiments revealed significant performance degradation in cross-lingual settings, particularly for reviews with complex emotional expressions or implicit semantics.
\citet{meng2025large} further found that current LLMs perform comparably to humans in identifying fluent fake reviews, highlighting the practical challenges faced by prompt-driven methods in deployment.

To improve detection performance, some studies propose guiding LLMs via prompts to generate explanatory features that are subsequently used for discriminative analysis.
\citet{Wang2026LLMIC} introduced the LLMIC framework, which employs LLaMA2-7B as a generator to produce extended texts and semantic explanations through prompts, followed by implicit feature extraction for classification.
On Amazon and hotel review datasets, this approach achieved accuracy rates above 93\%, with an F1 score of 95.9\% on the hotel dataset, outperforming conventional text embedding methods.
\citet{Liyanage2024AIEnhanced} applied GPT-based models to generate review summaries and argumentative structures on opinion spam datasets, using them as auxiliary features for classification.
Their results demonstrated that explanation-oriented prompts effectively enhance the detection of logical coherence and factual consistency.
\citet{yu2024dpicdecouplingpromptintrinsic} further provided a theoretical analysis of the decoupling between prompts and generated text features in the DPIC framework, proposing prompt reconstruction techniques to extract intrinsic statistical properties of LLM-generated text, thereby offering methodological support for explanation-driven detection with enhanced interpretability.

Prompt-driven data augmentation is another active research direction.
\citet{SALMINEN2022102771} were among the first to use GPT-2 to generate fake reviews for training, demonstrating that synthetic data can substantially improve the generalization ability of detection models.
\citet{liu2025data} subsequently used GPT-4 to generate cross-domain fake reviews and constructed multi-domain datasets spanning Amazon, Yelp, and DianPing, with experimental results showing accuracy improvements exceeding 12\% in low-resource scenarios.
\citet{liu2025detectingllmgeneratedspamreviews} proposed the FraudSquad framework, which uses prompts to control LLMs to generate diverse fake reviews and integrates graph neural networks to model user and review relationships, achieving detection accuracy ranging from 89.5\% to 99.9\% across three synthetic datasets with only 1\% labeled data.
\citet{gambetti2024aigenfoodreviewmultimodaldatasetmachinegenerated} further extended prompt-based augmentation to multimodal settings by introducing the AiGen-FoodReview dataset, which includes jointly generated text and image reviews, providing a valuable benchmark for multimodal fake review detection.

Given that the effectiveness of prompt-driven methods is highly sensitive to prompt design, recent studies have explored strategies that integrate prompts with model architectures.
\citet{Singh_2023} proposed a Stacked Transformer Ensemble approach that combines GPT-generated features with BERT-based embeddings via stacking, achieving higher detection accuracy than single-model baselines across multiple platform datasets.
\citet{Wu2025LLMFA} proposed optimizing prompt structures using evolutionary algorithms, automatically searching for effective instruction templates to enhance model sensitivity to semantic manipulation.

Prompt-driven methods offer clear advantages in annotation efficiency, cross-lingual transferability, and detection transparency through explanation generation.
However, when LLMs are used solely as final judges, their discriminative capability remains insufficient for reliable fake review detection, particularly against semantically sophisticated forgeries.
This limitation has motivated a shift toward treating LLMs as semantic representation and decision-enhancement modules.

\subsection{LLM-based Feature Extraction and Semantic Enhancement Methods}

Beyond direct classification through prompts, recent work uses LLMs as semantic representation extractors or discriminative enhancement modules within fake review detection frameworks.
This line of research treats LLMs as components for semantic modeling, representation learning, and decision enhancement rather than as end-to-end decision-makers.
Some studies discussed in this subsection come from adjacent tasks such as misinformation detection, AI-generated text detection, or general safety-related text classification.
They are included only when their mechanisms are directly relevant to fake review detection, such as semantic embedding fusion, graph-based propagation, contrastive representation learning, or paraphrase-robust detection.
These studies should be read as methodological evidence rather than direct empirical evidence on fake review benchmarks.
They are not counted in the coded corpus of 211 studies unless they include fake review detection experiments.
Accordingly, direct fake review studies are discussed first, while adjacent-task studies are used only to clarify transfer mechanisms that still require validation on fake review datasets.

Direct fake review studies increasingly use LLMs as auxiliary feature generators, semantic explainers, or representation modules rather than standalone judges.
\citet{Wang2026LLMIC} used LLaMA-based generation to derive implicit semantic characteristics for fake review classification.
\citet{liu2025detectingllmgeneratedspamreviews} combined LLM-generated review representations with heterogeneous graph modeling to connect semantic and behavioral evidence.
Adjacent fake news studies further show that LLMs can support advisory detection and fine-grained semantic analysis~\cite{hu2024bad,repede2024llama}, but these results are treated here only as methodological analogues.

A representative line of fake review research treats contextual embeddings produced by LLMs as semantic features that can be fused with behavioral or structural signals.
The adjacent-task evidence helps explain why this design is plausible, but it is not used here as a substitute for fake-review-specific evaluation.
\citet{Papageorgiou2025HarnessingLL} compared pretrained language models such as BERT and RoBERTa in misinformation detection and reported gains over bag-of-words models and static word embeddings for semantically ambiguous texts.
\citet{Ma2024LLMSemantics} further argued that a single text embedding is insufficient for deeper deceptive-content analysis and proposed integrating LLM-based semantic embeddings with topic information and entity relations through heterogeneous semantic graphs.
These results motivate joint semantic and structural modeling, while the fake review setting still requires validation against review-specific labels, reviewer histories, and platform-derived signals.

At a mechanism level, AI-generated text detection also offers useful evidence about paraphrase resistance and semantic similarity modeling.
\cite{he2024sefdsemanticenhancedframeworkdetecting} introduced the Semantic-Enhanced Fake Detection framework, which employs pretrained language models such as BERT and Sentence-BERT to extract text embeddings and combines similarity modeling with downstream classifiers.
This framework demonstrated stable performance across various AI-generated text detection tasks, showing particular robustness against paraphrasing attacks.
To further improve the detection of complex and coordinated deceptive behaviors, some studies integrate LLM embeddings with graph neural networks.
\citet{liu2025detectingllmgeneratedspamreviews} constructed heterogeneous graphs by combining LLM-based review embeddings with user and review interaction structures, enabling joint modeling of semantic and behavioral signals.
Experimental results across multiple LLM-generated fake review datasets showed marked improvements in Precision, Recall, and F1 scores compared with text-only models.
These findings indicate that LLM embeddings can function not only as standalone features but also as node representations propagated within graph structures, facilitating the detection of collective fraud patterns and coordinated semantic manipulation.

Contrastive learning has been adopted to optimize semantic embedding spaces by increasing the separability between genuine and fake texts.
\citet{sanchez2025semanticawarecontrastivefinetuningboosting} proposed a semantic-aware contrastive fine-tuning strategy that constructs positive and negative sample pairs to reshape embedding distributions, making authentic and generated texts more distinguishable in high-dimensional semantic spaces.
Although primarily applied to other safety-related detection tasks, the underlying principle of embedding-level discriminative optimization is relevant to fake review detection through semantic representation adaptation.
From the perspectives of interpretability and robustness, \citet{wu2025selfregularizationsparseautoencoderscontrollable} introduced latent-space self-regularization mechanisms to constrain and explain LLM embeddings, guiding models to focus on task-relevant semantic dimensions with greater precision.
Their experiments demonstrated improved generalization performance and reduced sensitivity to noisy features across multiple text classification tasks under diverse settings in practical scenarios.

Beyond traditional encoding-based embeddings, some studies explore generative approaches to further enhance semantic representation quality.
\citet{tsai2025letllmsspeakembedding} proposed an iterative generation and contrastive semantic embedding framework, which alternates between generation and contrastive learning to jointly optimize embedding stability and semantic discriminability.
Although primarily designed for semantic similarity modeling, its generative embedding paradigm offers valuable insights for semantic discrimination in fake review detection.

In practical applications, the fusion of multiple feature sources improves detection performance when the sources provide complementary evidence.
\citet{arshed2025contextaware} developed a context-aware representation learning framework that integrates LLM semantic embeddings with stylistic and statistical features to distinguish human-authored text from AI-generated content, achieving F1 scores exceeding 96\% on multi-domain tweet datasets.
Several studies adopt hybrid neural architectures that combine BERT-style embeddings with convolutional networks for local pattern extraction and handcrafted statistical features, as demonstrated by \citet{Alharthi2025} and \citet{AlJamal2025}, yielding stable and reproducible performance gains in generated text detection tasks.

Taken together, this evidence suggests that LLMs are often more useful as semantic representation and discriminative enhancement modules than as standalone prompt-driven judges.
LLM embeddings can capture implicit meaning, discourse coherence, and argumentative structure~\cite{Papageorgiou2025HarnessingLL,Ma2024LLMSemantics}.
Their modular nature also allows integration with graph neural networks, contrastive learning objectives, and multimodal architectures~\cite{liu2025detectingllmgeneratedspamreviews,sanchez2025semanticawarecontrastivefinetuningboosting}.
Evidence from multi-domain and multilingual settings further suggests potential generalization benefits~\cite{he2024sefdsemanticenhancedframeworkdetecting,arshed2025contextaware}.
Nevertheless, the computational and storage costs associated with large-scale embedding extraction remain substantial, and embedding-based decision boundaries can be sensitive to distributional shifts across tasks and datasets.
The continued reliance on synthetic training data in many studies also raises questions about generalization to deployment conditions, motivating further investigation into resilient and data-efficient representation learning strategies.

\subsection{Generative Discriminative Methods from GANs to LLMs}

Generative discriminative approaches have gradually emerged as an important research direction in fake review detection.
The core idea is to characterize the distributional differences between genuine and deceptive reviews by constructing collaborative mechanisms between generative models and discriminative models.
Such methods typically incorporate techniques such as Generative Adversarial Networks, language model likelihood estimation, and self-consistency mechanisms to enhance the detection of fluent fake reviews.
Unlike traditional discriminative classifiers that primarily focus on learning decision boundaries, generative discriminative methods emphasize explicit modeling of the latent distribution of review texts through generation processes.
This design improves robustness in challenging scenarios such as semantic paraphrasing, sentiment camouflage, and reviews generated by advanced LLMs.

Traditional GANs and autoregressive LLMs are treated as separate model families in this survey.
FakeGAN, ScoreGAN, and semi-supervised GANs are discussed here as predecessors of generation-aware detection, but they remain classified as DL-based methods in the taxonomy and performance tables.
The later part of this subsection covers methods that use autoregressive language models for likelihood estimation, review generation, or multi-stage discrimination.

Generative Adversarial Networks were first introduced in text generation tasks and were subsequently extended to the domain of fake review detection.
\citet{aghakhani2018detecting} proposed the FakeGAN framework, which constructs an adversarial learning architecture composed of a generator and two discriminators.
In this framework, the generator continuously produces deceptive fake review samples, thereby forcing the discriminators to learn more resilient decision boundaries.
Their results demonstrate that adversarially generated samples can improve the model's ability to identify covert fake reviews.
Building upon this idea, \citet{shehnepoor2021scoreganfraudreviewdetector} proposed ScoreGAN, which jointly models rating information and textual content.
By adopting a multi-task learning strategy, the training process of both the generator and discriminator is constrained to capture the interaction patterns between review texts and rating manipulation behaviors.
Experimental results show that this method outperforms traditional deep learning models on the Yelp and TripAdvisor datasets.

To address the scarcity of labeled fake review data, several studies have extended adversarial generation into semi-supervised learning frameworks.
In this setting, the generator is used to synthesize fake review samples, while the discriminator is jointly responsible for authenticity discrimination and category learning, thereby improving generalization performance under limited labeled data.
\citet{shawon2023bengali} proposed a semi-supervised GAN-based fake review detection model built upon BanglaBERT for low-resource language settings.
By expanding the training sample space through the generator, the discriminator is able to learn richer language distributional features.
Experimental results on Bangla product review datasets show that this approach outperforms supervised methods relying solely on pretrained language models, validating the effectiveness of generative discriminative strategies in low-resource environments.

Beyond adversarial learning, some studies directly use the probabilistic outputs of generative models as discriminative signals.
These approaches typically rely on the generation likelihood or cumulative probability density of language models to determine whether a review deviates from the distribution of genuine human-written text.
\citet{Luo2023AIGeneratedFakeReview} proposed a fake review detection method based on cumulative probability density, which identifies anomalous texts generated by language models by computing tail probabilities under the language model distribution.
This method does not require explicit adversarial training, but instead characterizes generation traces from a statistical distribution perspective, exhibiting good stability and interpretability in AI-generated review detection scenarios.

With the rapid advancement of LLMs, several studies have proposed multi-stage generative discriminative frameworks.
In such methods, the first stage employs generative models to produce fluent machine-generated reviews, while the second stage integrates multiple discriminative models spanning convolutional networks, bidirectional recurrent architectures, and Transformer-based classifiers.
\citet{Nawara2025DualPhaseLLM} provide empirical evidence that multi-stage generation and discrimination strategies can effectively cover diverse types of generative forgery behaviors and maintain consistently high detection accuracy even when machine-generated reviews closely resemble genuine ones.

Generative discriminative approaches introduce generation mechanisms, probabilistic modeling, and adversarial learning into fake review detection, offering a perspective that extends beyond purely discriminative models.
These methods face persistent challenges including training instability, high computational costs, and the absence of widely accepted unified evaluation standards.
The following subsection examines how LLM-based data augmentation and adversarial learning can further address the data scarcity and class imbalance challenges that constrain the practical deployment of fake review detection systems.

\subsection{LLM-based Data Augmentation and Adversarial Learning Methods}

In fake review detection tasks, data scarcity and class imbalance limit model performance.
LLM-based data augmentation and adversarial learning therefore aim to improve detection under constrained labeling conditions.
The core idea of data augmentation is to expand the training set using generative models or other strategies, improving the model's ability to learn from minority classes or semantically complex samples with greater representation diversity.
\citet{liu2025data} proposed cross-lingual and cross-domain data augmentation methods using LLMs such as GPT-2 and OPT to generate diverse review samples, which are then combined with original datasets for training.
Experiments demonstrate that on datasets such as DeRev, Amazon, Yelp, and DianPing, augmented data can increase detection accuracy by approximately 0.3 to 10.9 percentage points, mitigating data insufficiency.
Their further investigation shows that the style of generated data is significantly influenced by the quality of prompts, and that when real fake reviews are used as generation seeds, the resulting synthetic reviews more closely resemble genuine deceptive reviews, thereby enhancing detection generalization.
\citet{SALMINEN2022102771} were among the first to use GPT-2 to generate fake reviews for training classifiers, marking an early application of data augmentation in fake review detection that increased the size of the training set.
\cite{Geetha2025MBODeBERTa} proposed a pretrained model-based augmentation technique that expands small original datasets multiple times while maintaining semantic consistency and diversity, enabling detectors to better capture semantic differences in complex deployment scenarios.

Adversarial learning aims to guide the model toward more resilient discriminative boundaries by generating perturbed or hard-to-classify samples.
In fake review detection, such methods typically involve adversarial sample generation, adversarial perturbation training, and multi-task adversarial learning.
\citet{shehnepoor2021scoreganfraudreviewdetector} proposed ScoreGAN, which jointly models review text and rating information to generate rating-related fake reviews, augmenting the discriminator's training set and thereby improving detection performance.
Another line of work employs sentiment-enhanced BERT within a multi-task adversarial learning framework, incorporating sentiment features as adversarial signals so that the discriminator learns richer semantic and emotional patterns.
For low-resource languages and tasks with insufficient labeled data, semi-supervised adversarial learning approaches that combine GAN frameworks with pretrained language models have been proposed to construct more challenging adversarial training samples, enhancing discriminator robustness~\cite{shawon2023bengali}.

With the advancement of adversarial learning theory, researchers have explored diversified mechanisms for adversarial sample generation, including semantic perturbations, style transformations, and structural rewrites, to produce reviews that are more difficult to detect and that force detection models to learn more abstract discriminative features.
These adversarial samples can be generated automatically using LLMs or combined with rule-based and style-based transformations to increase semantic diversity.
Prior work on fake news has demonstrated the effectiveness of style-based adversarial augmentation strategies, in which LLM-generated style variants are used to train detectors for increased robustness against varied falsification strategies.
Similar augmentation logic can inform fake review detection, but the transfer should be verified on review-specific benchmarks~\cite{tong2025generate}.
Across different datasets, LLM-generated augmented data often improves performance, with models trained on augmented data showing gains in accuracy and recall compared to models trained on original data alone~\cite{liu2025data}.
GAN-based adversarial sample methods additionally alleviate class imbalance by sharpening the discriminator's decision boundary across different semantic ranges of fake reviews.

The use of LLM-generated fake reviews as training resources has also been extensively studied.
\citet{liu2025data} proposed using LLMs including GPT and GLM to generate synthetic fake reviews for training set expansion, enhancing model generalization across domains, languages, and platforms.
\citet{tong2025generate} explored adversarial generation with collaborative optimization by combining reinforcement learning with LLMs to produce diverse and covert fake content.
\citet{qian2025glm} systematically analyzed GLM-based models in multi-domain synthetic information generation, demonstrating their ability to produce highly realistic and coherent review texts and providing a technical foundation for large-scale automated fake review generation.
\citet{bader2023detecting} conducted a supervised analysis of ChatGPT-generated fake reviews, finding that while language fluency and sentiment expression are close to human-written texts, deviations remain in emotional fluctuation, personalization, and repetition patterns, which can serve as cues for detection.
\citet{sadasivan2023can} noted that traditional detection methods relying on feature distribution differences exhibit reduced accuracy against LLM-generated texts, highlighting limitations in generalization and robustness.
\citet{meng2025large} observed that LLM-generated reviews, while highly realistic in grammar and sentiment, may possess enhanced persuasiveness, effectively misleading consumer judgment and amplifying the social impact of fake reviews.
\citet{wu2025survey} summarized current technological pathways into three categories covering probability feature-based methods, model comparison-based methods, and watermark and statistical detection-based methods, providing a methodological framework for constructing interpretable review detection systems.
\citet{rao2025detecting} proposed a hybrid method combining watermarking with statistical detection for identifying LLM-generated peer review texts, a strategy applicable to high-similarity review scenarios.

Augmented data allow detectors to learn effective discriminative features even under scarce labeled conditions, as LLM-generated synthetic reviews contain diverse semantic patterns that mitigate overfitting to single distributions.
Adversarial samples and mixed training mechanisms improve model sensitivity to difficult-to-distinguish forgery strategies.
The quality of augmented data, however, depends heavily on prompt design and generative model performance, introducing risks of generation bias and noise, while poorly designed perturbation strategies in adversarial learning may produce misleading training signals.

\subsection{Limitations and Synthesis}

LLM-based fake review detection currently spans prompt-driven direct discrimination, semantic representation and discriminative enhancement, generative discrimination, and data augmentation with adversarial learning.
Across these categories, the field is moving from treating LLMs as black-box discriminators toward embedding them as components within multi-source detection frameworks that integrate textual, behavioral, structural, and external knowledge signals.

LLMs occupy a complex and evolving dual role in the fake review ecosystem.
Their generative capabilities have substantially raised the sophistication of fabricated reviews, producing content that closely mimics authentic user writing in fluency, sentiment, and discourse structure, thereby rendering many conventional detection signals unreliable.
At the same time, their semantic reasoning capabilities, when properly integrated with graph-based modeling, contrastive learning, and knowledge-enhanced architectures, provide detection systems with a depth of understanding that shallow feature-based methods cannot achieve.
The co-evolution of generation and detection defines the central challenge of this field and calls for detection frameworks that reason across multiple sources of evidence.

From the perspective of information fusion, the most effective approaches reviewed in this section are those that move beyond single-modality text analysis to incorporate heterogeneous signals, including reviewer behavior sequences, product-user interaction graphs, external knowledge bases, and multimodal content features.
The convergence of LLM-based semantic modeling with structured relational reasoning combines large-scale language understanding with graph-based structural constraints.

Despite these advances, several limitations still persist across the reviewed methods. The interpretability of LLM-based detection decisions remains limited, and the potential for hallucinated reasoning in chain-of-thought prompting introduces significant reliability concerns in high-stakes deployment scenarios. Cross-domain generalization continues to degrade under distributional shift, and the dependence on synthetic training data raises questions about ecological validity. These challenges motivate the research directions discussed in the following section.

\section{Comparative Analysis and Practical Insights}
\label{sec:Comparative}
\subsection{Performance Evolution on Representative Benchmarks}
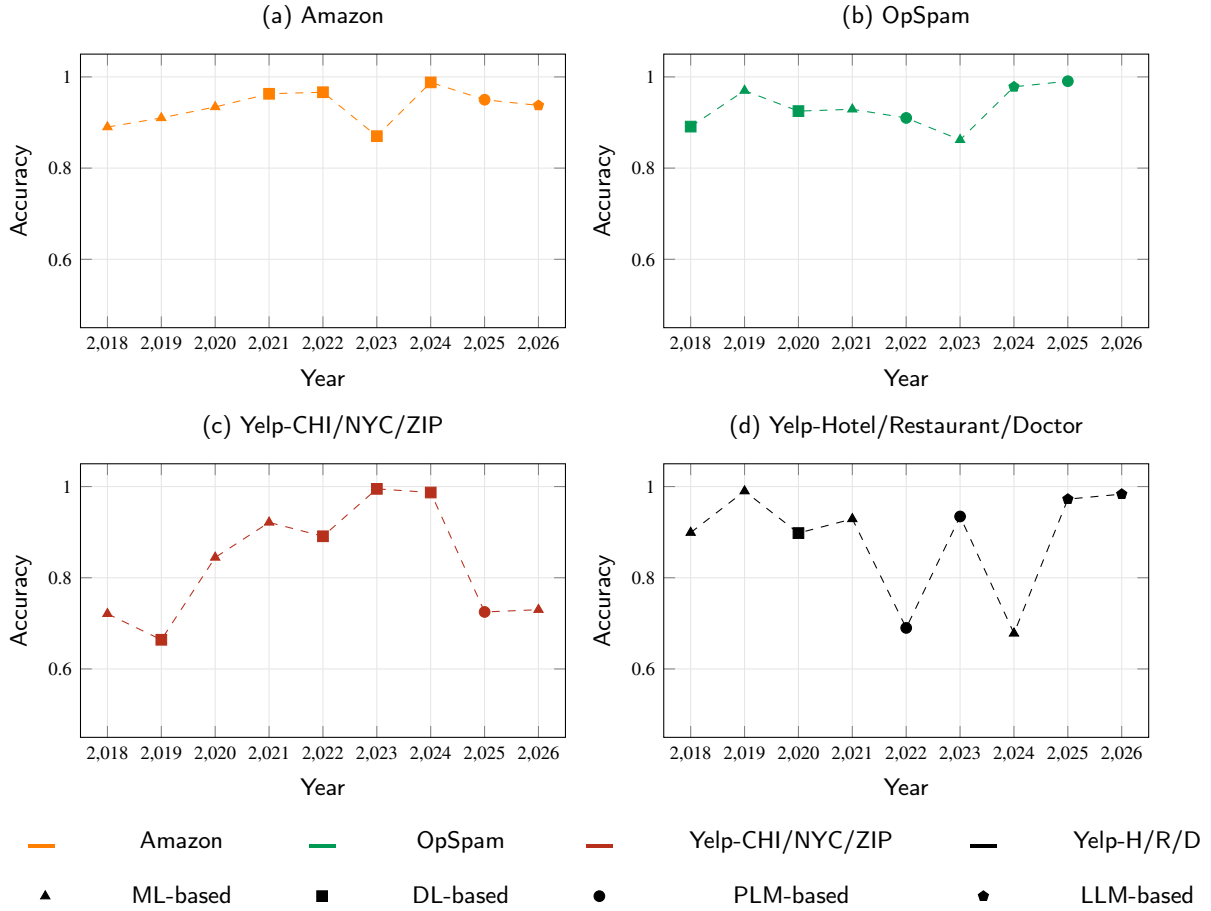
\begin{figure*}[t]
\centering
\begin{tikzpicture}

\newcommand{\trianglemark}{
\tikz[baseline=-0.6ex]{
\draw[mark=triangle*,mark options={solid}]
plot coordinates {(0,0)};
}
}
\newcommand{\squaremark}{
\tikz[baseline=-0.6ex]{
\draw[mark=square*,mark options={solid}]
plot coordinates {(0,0)};
}
}
\newcommand{\circlemark}{
\tikz[baseline=-0.6ex]{
\draw[mark=*,mark options={solid}]
plot coordinates {(0,0)};
}
}
\newcommand{\pentagonmark}{
\tikz[baseline=-0.6ex]{
\draw[mark=pentagon*,mark options={solid}]
plot coordinates {(0,0)};
}
}
\begin{groupplot}[
group style={
group size=2 by 2,
horizontal sep=1.3cm,
vertical sep=1.8cm
},
width=0.46\textwidth,
height=5.2cm,
xmin=2017.5,
xmax=2026.5,
ymin=0.45,
ymax=1.05,
xtick={2018,...,2026},
grid=major,
grid style={gray!20},
tick label style={font=\scriptsize},
label style={font=\small},
title style={font=\small},
xlabel={Year},
ylabel={Accuracy}
]

\nextgroupplot[title={(a) Amazon}]

\addplot[orange,dashed]
coordinates{
(2018,0.89)
(2019,0.91)
(2020,0.9341)
(2021,0.963)
(2022,0.9664)
(2023,0.87)
(2024,0.988)
(2025,0.95)
(2026,0.9375)
};

\addplot[
only marks,
mark=triangle*,
orange]
coordinates{
(2018,0.89)
(2019,0.91)
(2020,0.9341)
};

\addplot[
only marks,
mark=square*,
orange]
coordinates{
(2021,0.963)
(2022,0.9664)
(2023,0.87)
(2024,0.988)
};

\addplot[
only marks,
mark=*,
orange]
coordinates{
(2025,0.95)
};

\addplot[
only marks,
mark=pentagon*,
mark options={solid},
orange]
coordinates{
(2026,0.9375)
};

\nextgroupplot[title={(b) OpSpam}]

\addplot[ForestGreen,dashed]
coordinates{
(2018,0.891)
(2019,0.97)
(2020,0.925)
(2021,0.929)
(2022,0.91)
(2023,0.862)
(2024,0.9783)
(2025,0.9906)
};

\addplot[
only marks,
mark=triangle*,
ForestGreen]
coordinates{
(2019,0.97)
(2021,0.929)
(2023,0.862)
};

\addplot[
only marks,
mark=square*,
ForestGreen]
coordinates{
(2018,0.891)
(2020,0.925)
};

\addplot[
only marks,
mark=*,
ForestGreen]
coordinates{
(2022,0.91)
(2025,0.9906)
};

\addplot[
only marks,
mark=pentagon*,
mark options={solid},
ForestGreen]
coordinates{
(2024,0.9783)
};

\nextgroupplot[title={(c) Yelp-CHI/NYC/ZIP}]

\addplot[BrickRed,dashed]
coordinates{
(2018,0.721)
(2019,0.664)
(2020,0.8445)
(2021,0.9214)
(2022,0.891)
(2023,0.995)
(2024,0.987)
(2025,0.725)
(2026,0.73)
};

\addplot[
only marks,
mark=triangle*,
BrickRed]
coordinates{
(2018,0.721)
(2020,0.8445)
(2021,0.9214)
(2026,0.73)
};

\addplot[
only marks,
mark=square*,
BrickRed]
coordinates{
(2019,0.664)
(2022,0.891)
(2023,0.995)
(2024,0.987)
};

\addplot[
only marks,
mark=*,
BrickRed]
coordinates{
(2025,0.725)
};

\nextgroupplot[
title={(d) Yelp-Hotel/Restaurant/Doctor}
]

\addplot[black,dashed]
coordinates{
(2018,0.899)
(2019,0.99)
(2020,0.898)
(2021,0.929)
(2022,0.69)
(2023,0.9344)
(2024,0.678)
(2025,0.9725)
(2026,0.9834)
};

\addplot[
only marks,
mark=triangle*,
black]
coordinates{
(2018,0.899)
(2019,0.99)
(2021,0.929)
(2024,0.678)
};

\addplot[
only marks,
mark=square*,
black]
coordinates{
(2020,0.898)
};

\addplot[
only marks,
mark=*,
black]
coordinates{
(2022,0.69)
(2023,0.9344)
};

\addplot[
only marks,
mark=pentagon*,
mark options={solid},
black]
coordinates{
(2025,0.9725)
(2026,0.9834)
};

\end{groupplot}

\matrix[
draw=none,
matrix of nodes,
column sep=8mm,
row sep=2mm,
anchor=north,
font=\small]
at ($(group c1r2.south)!0.5!(group c2r2.south)+(0,-1.0cm)$)
{

\textcolor{orange}{\rule{10pt}{1.2pt}}
& Amazon &
\textcolor{ForestGreen}{\rule{10pt}{1.2pt}}
& OpSpam &
\textcolor{BrickRed}{\rule{10pt}{1.2pt}}
& Yelp-CHI/NYC/ZIP &
\textcolor{black}{\rule{10pt}{1.2pt}}
& Yelp-H/R/D
\\

\trianglemark
& ML-based &
\squaremark
& DL-based &
\circlemark
& PLM-based &
\pentagonmark
& LLM-based
\\
};
\end{tikzpicture}
\caption{Accuracy trend on representative fake review detection benchmarks.}
\label{fig:acc_performance_evolution}
\end{figure*}

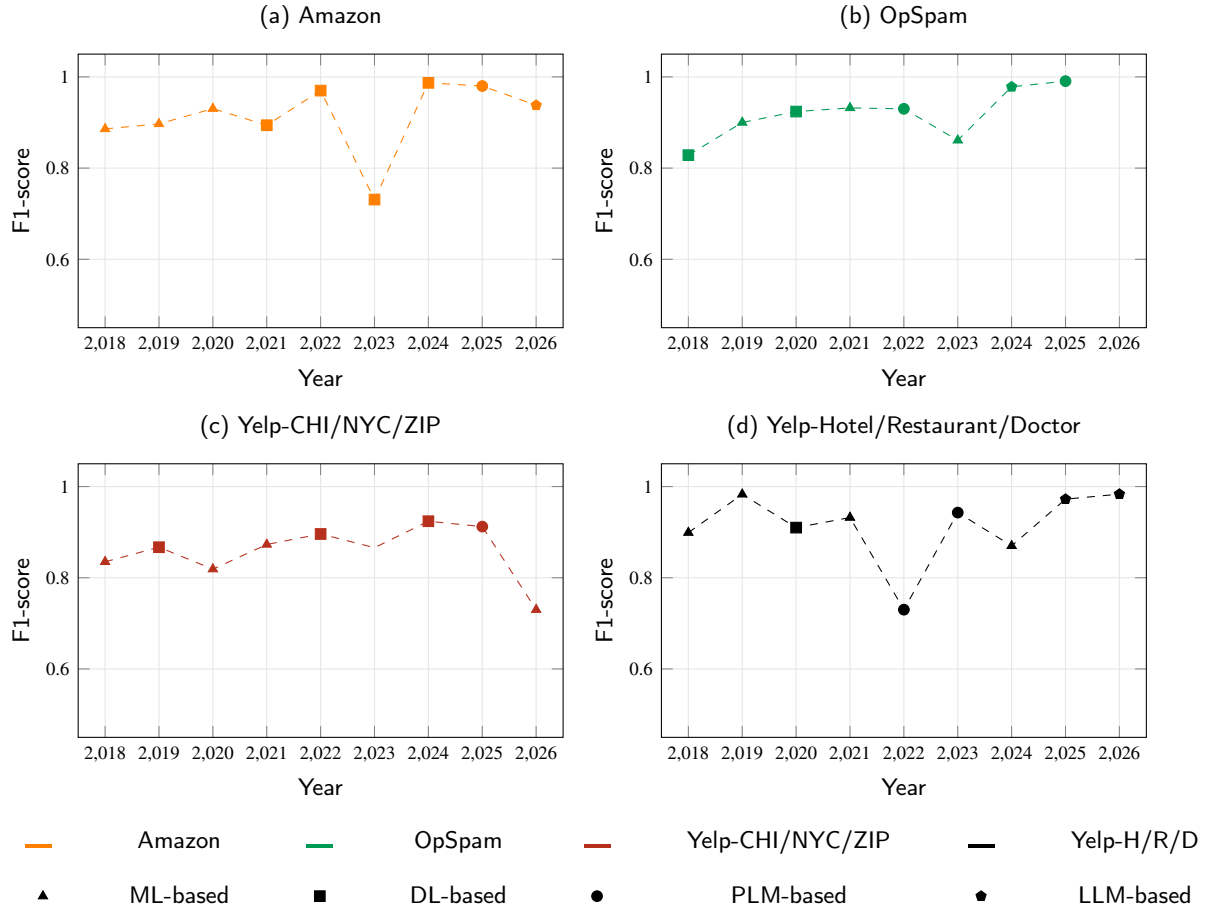
\begin{figure*}[t]
\centering
\begin{tikzpicture}

\newcommand{\trianglemark}{
\tikz[baseline=-0.6ex]{
\draw[mark=triangle*,mark options={solid}]
plot coordinates {(0,0)};
}
}

\newcommand{\squaremark}{
\tikz[baseline=-0.6ex]{
\draw[mark=square*,mark options={solid}]
plot coordinates {(0,0)};
}
}

\newcommand{\circlemark}{
\tikz[baseline=-0.6ex]{
\draw[mark=*,mark options={solid}]
plot coordinates {(0,0)};
}
}

\newcommand{\pentagonmark}{
\tikz[baseline=-0.6ex]{
\draw[mark=pentagon*,mark options={solid}]
plot coordinates {(0,0)};
}
}

\begin{groupplot}[
group style={group size=2 by 2,horizontal sep=1.3cm,vertical sep=1.8cm},
width=0.46\textwidth,
height=5.2cm,
xmin=2017.5,
xmax=2026.5,
ymin=0.45,
ymax=1.05,
xtick={2018,...,2026},
grid=major,
grid style={gray!20},
tick label style={font=\scriptsize},
label style={font=\small},
title style={font=\small},
xlabel={Year},
ylabel={F1-score}
]

\nextgroupplot[title={(a) Amazon}]

\addplot[orange,dashed]
coordinates{
(2018,0.886)
(2019,0.897)
(2020,0.9304)
(2021,0.894)
(2022,0.970)
(2023,0.731)
(2024,0.987)
(2025,0.980)
(2026,0.9378)
};

\addplot[only marks,mark=triangle*,orange]
coordinates{
(2018,0.886)
(2019,0.897)
(2020,0.9304)
};

\addplot[only marks,mark=square*,orange]
coordinates{
(2021,0.894)
(2022,0.970)
(2023,0.731)
(2024,0.987)
};

\addplot[only marks,mark=*,orange]
coordinates{
(2025,0.980)
};

\addplot[only marks,mark=pentagon*,mark options={solid},orange]
coordinates{
(2026,0.9378)
};

\nextgroupplot[title={(b) OpSpam}]

\addplot[ForestGreen,dashed]
coordinates{
(2018,0.8284)
(2019,0.9000)
(2020,0.9240)
(2021,0.9320)
(2022,0.9300)
(2023,0.8610)
(2024,0.9783)
(2025,0.9907)
};

\addplot[only marks,mark=triangle*,ForestGreen]
coordinates{
(2019,0.9000)
(2021,0.9320)
(2023,0.8610)
};

\addplot[only marks,mark=square*,ForestGreen]
coordinates{
(2018,0.8284)
(2020,0.9240)
};

\addplot[only marks,mark=*,ForestGreen]
coordinates{
(2022,0.9300)
(2025,0.9907)
};

\addplot[only marks,mark=pentagon*,mark options={solid},ForestGreen]
coordinates{
(2024,0.9783)
};

\nextgroupplot[title={(c) Yelp-CHI/NYC/ZIP}]

\addplot[BrickRed,dashed]
coordinates{
(2018,0.8350)
(2019,0.8670)
(2020,0.8189)
(2021,0.8730)
(2022,0.8960)
(2023,0.8660)
(2024,0.9240)
(2025,0.9122)
(2026,0.7300)
};

\addplot[only marks,mark=triangle*,BrickRed]
coordinates{
(2018,0.8350)
(2020,0.8189)
(2021,0.8730)
(2026,0.7300)
};

\addplot[only marks,mark=square*,BrickRed]
coordinates{
(2019,0.8670)
(2022,0.8960)
(2024,0.9240)
};

\addplot[only marks,mark=*,BrickRed]
coordinates{
(2025,0.9122)
};

\nextgroupplot[title={(d) Yelp-Hotel/Restaurant/Doctor}]

\addplot[black,dashed]
coordinates{
(2018,0.8990)
(2019,0.9830)
(2020,0.9100)
(2021,0.9320)
(2022,0.7300)
(2023,0.9429)
(2024,0.8700)
(2025,0.9724)
(2026,0.9834)
};

\addplot[only marks,mark=triangle*,black]
coordinates{
(2018,0.8990)
(2019,0.9830)
(2021,0.9320)
(2024,0.8700)
};

\addplot[only marks,mark=square*,black]
coordinates{
(2020,0.9100)
};

\addplot[only marks,mark=*,black]
coordinates{
(2022,0.7300)
(2023,0.9429)
};

\addplot[only marks,mark=pentagon*,mark options={solid},black]
coordinates{
(2025,0.9724)
(2026,0.9834)
};

\end{groupplot}

\matrix[
draw=none,
matrix of nodes,
column sep=8mm,
row sep=2mm,
anchor=north,
font=\small]
at ($(group c1r2.south)!0.5!(group c2r2.south)+(0,-1.0cm)$)
{
\textcolor{orange}{\rule{10pt}{1.2pt}} & Amazon &
\textcolor{ForestGreen}{\rule{10pt}{1.2pt}} & OpSpam &
\textcolor{BrickRed}{\rule{10pt}{1.2pt}} & Yelp-CHI/NYC/ZIP &
\textcolor{black}{\rule{10pt}{1.2pt}} & Yelp-H/R/D
\\

\trianglemark & ML-based &
\squaremark & DL-based &
\circlemark & PLM-based &
\pentagonmark & LLM-based
\\
};
\end{tikzpicture}
\caption{F1-score trend on representative fake review detection benchmarks.}
\label{fig:f1_performance_evolution}
\end{figure*}

To investigate the evolution of fake review detection approaches beyond isolated performance reports, we analyze historical performance changes of representative methods on four widely used benchmark families, including Amazon, OpSpam, Yelp-CHI/NYC/ZIP, and Yelp-Hotel/Restaurant/Doctor.
Figure~\ref{fig:acc_performance_evolution} and Figure~\ref{fig:f1_performance_evolution} present the evolution of Accuracy and F1-score from 2018 to 2026.
Different marker shapes indicate detection paradigms, including ML-based, DL-based, PLM-based, and LLM-based methods, while different colors represent benchmark datasets.
These curves should be interpreted as descriptive trends rather than controlled head-to-head rankings.
The reported results come from studies with different data splits, label construction procedures, preprocessing strategies, and evaluation protocols.
Accordingly, the analysis emphasizes broad shifts in model families and evidence fusion strategies rather than claiming strict superiority of individual methods.
The plotted values were manually extracted from the representative method tables in this survey and aligned by publication year, benchmark family, dominant modeling family, and reported metric.
When a study reported multiple configurations on the same benchmark, the plotted point reflects the reported configuration used by the original authors to summarize that method.
Each plotted point corresponds to a method row in Table~\ref{tab:ml_results_3}, Table~\ref{tab:dl_results_1}, Table~\ref{tab:dl_results_2}, Table~\ref{tab:plm_results}, or Table~\ref{tab:llm_results}.
Rows without the relevant reported metric were not plotted.
The figure therefore supports a traceable overview of reported progress, but it does not estimate an average effect size or a statistically controlled benchmark ranking.

\textbf{Evolution on Amazon Benchmarks.} Reported Amazon results show a broad upward pattern in both Accuracy and F1-score, but the points should not be read as a controlled time series. Early ML-based methods reported moderate performance, with Accuracy values ranging from approximately 0.89 in 2018 to 0.9341 in 2020. These values suggest that handcrafted linguistic and behavioral features can provide useful signals for identifying deceptive reviews. Later DL-based reports reached approximately 0.97 F1-score and 0.9664 Accuracy in 2022, which is consistent with the stronger contextual modeling capacity of neural architectures for fake review detection.

The reported Amazon results also show non-monotonic variation. Around 2023, some reported configurations show lower Accuracy and F1-score than nearby years. Such variation is more likely to reflect differences in label construction, preprocessing, model design, and data splits than a true year-by-year decline. Recent PLM- and LLM-based studies report competitive values, with Accuracy close to or above 0.95 and F1-score approaching 0.98 in some settings. These results suggest possible benchmark saturation, but this interpretation should be treated cautiously because the underlying experimental protocols differ across studies.

\textbf{Evolution on OpSpam Benchmarks.}
Compared with Amazon, OpSpam demonstrates consistently high performance across different generations of methods. Both Accuracy and F1-score remain above 0.85 for most years in the extracted records, suggesting that this dataset contains more relatively distinguishable deceptive patterns. ML-based methods already reported strong performance, with Accuracy reaching 0.97 in 2019 and F1-score exceeding 0.90. This pattern suggests that feature-based approaches remain highly effective when deceptive cues are explicit and linguistically consistent across domains.

DL-based methods maintained reported F1 scores around 0.92 to 0.93. Later PLM- and LLM-based studies reported values close to 0.99 in some configurations. The limited margin between older and newer reports suggests that OpSpam may have limited discriminative power for evaluating advanced foundation models, although this conclusion remains constrained by protocol differences.

Therefore, although OpSpam remains one of the most frequently used benchmarks, its near-saturated performance distribution highlights the necessity of developing more challenging datasets containing diverse domains, realistic adversarial behaviors, and LLM-generated deceptive reviews with greater ecological validity and complexity.

\textbf{Evolution on Yelp-CHI/NYC/ZIP Benchmarks.} The Yelp-CHI/NYC/ZIP benchmark presents a more variable reported performance pattern than Amazon and OpSpam. Accuracy and F1-score fluctuate across years, which is consistent with the greater diversity in Yelp review environments and behavioral signals. Early ML-based approaches reported Accuracy around 0.72 in 2018. Some later DL-based studies reported Accuracy close to 0.99, but these high values should be interpreted in relation to their specific split and label protocol carefully.

Interestingly, the F1-score trajectory does not completely follow the Accuracy trend. For example, some methods achieve high Accuracy but relatively lower F1-score, suggesting potential class imbalance issues or uneven detection capability between genuine and deceptive reviews. This observation indicates that Accuracy alone may provide an incomplete evaluation of fake review detection systems, especially when fraudulent reviews represent a minority class.

Recent PLM-based methods report competitive F1-score performance, but the gains are less consistent than those reported on Amazon or OpSpam. This pattern points to domain variation as a persistent challenge and supports more careful evaluation of cross-domain adaptation and behavioral information integration.

\textbf{Evolution on Yelp-Hotel/Restaurant/Doctor Benchmarks.}
The Yelp-Hotel/Restaurant/Doctor benchmark exhibits the largest variation among the four benchmark families. Early ML-based methods reported relatively high values, with Accuracy around 0.90 to 0.99 and F1-score above 0.90. Later approaches show considerable fluctuations in the extracted records. For example, some PLM-based reports around 2022 are lower than later reports.

Recent LLM-based approaches report strong results, reaching approximately 0.98 Accuracy and F1-score in some settings. These values suggest that large language models can capture subtle semantic inconsistencies and domain-specific deceptive patterns. However, the fluctuations observed across studies show that domain-specific characteristics remain a major challenge, particularly when models are transferred across review categories.
Across all benchmarks, several important observations can be derived from the performance evolution.

First, fake review detection has experienced a methodological transition from feature engineering-based ML methods toward representation learning and reasoning-oriented approaches. DL methods improved semantic feature extraction, while PLMs and LLMs further enhanced contextual understanding and generalization capability.
Second, Accuracy and F1-score exhibit similar but not identical evolutionary trends. Although Accuracy generally increases over time, F1-score reveals additional challenges caused by class imbalance and asymmetric detection performance. Therefore, future evaluation should consider multiple complementary metrics rather than relying solely on Accuracy.
Third, recent improvements from PLMs and LLMs appear incremental on traditional benchmarks. Existing datasets, particularly Amazon and OpSpam, may be approaching reported performance saturation, which limits their ability to distinguish between advanced detection models. Future benchmark construction should incorporate more realistic scenarios, including LLM-generated reviews, adversarial attacks, multimodal content, and cross-platform evaluation.

Finally, the evolution trend suggests that the next generation of fake review detection systems will move beyond text classification toward multi-source reasoning frameworks that integrate textual semantics, reviewer behaviors, temporal patterns, social relationships, and external knowledge.

\subsection{Comprehensive Comparison of Detection Paradigms}

Different fake review detection paradigms exhibit distinct advantages and limitations due to their underlying learning mechanisms and representation capabilities.
Traditional machine learning models mainly rely on handcrafted linguistic, behavioral, and statistical features, while deep learning approaches automatically learn hierarchical
representations from review content.
More recently, pretrained language models and large language models have introduced large-scale semantic knowledge and reasoning capabilities into fake review detection.

To provide practical guidance for model selection, we compare representative ML-based, DL-based, PLM-based, and LLM-based approaches from eight dimensions, including learning paradigm, feature representation, annotation requirement, domain adaptation capability, computational characteristics, interpretability, and generalization ability.
The comparison is summarized in Table~\ref{tab:paradigm_comparison}.

\begin{sidewaystable*}[p]

\centering

\caption{Comprehensive comparison of representative fake review detection models across different paradigms.}

\label{tab:paradigm_comparison}

\scriptsize

\begin{adjustbox}{max width=\textheight}
\begin{tabular}{p{1.8cm}p{1cm}p{1.2cm}p{2.2cm}p{2.5cm}p{2.8cm}p{2.5cm}p{2.5cm}p{3.0cm}}

\toprule

\textbf{Representative models}
&
\textbf{Ref.}
&
\textbf{Paradigm}
&
\textbf{Learning paradigm}
&
\textbf{Feature representation}
&
\textbf{Domain adaptation}
&
\textbf{Computational characteristics}
&
\textbf{Interpretability mechanism}
&
\textbf{Generalization ability}
\\

\midrule
SVM
&
\cite{ott2011finding,mukherjee2013whatyf,6920707}
&
ML-based
&
Supervised classification
&
TF-IDF, n-gram, LIWC, stylistic and psycholinguistic features
&
Limited due to domain-specific handcrafted features
&
Low training and inference cost
&
Feature weights and margin-based decision boundary
&
Good on same-domain datasets but limited cross-domain robustness
\\

LR
&
\cite{jindal2008opinion,rayana2015collective}
&
ML-based
&
Supervised probabilistic classification
&
Lexical features, rating deviation, reviewer behavioral statistics, metadata features
&
Limited, relies on manually designed domain-specific features
&
Low computational cost and efficient inference
&
Explicit feature coefficients provide interpretable contribution analysis
&
Strong on specific platforms but limited under cross-domain distribution shifts
\\

Random Forest
&
\cite{shan2021conflicts}
&
ML-based
&
Supervised ensemble learning based on multiple decision trees
&
Lexical features, sentiment indicators, reviewer behavior, rating patterns, metadata and relational features
&
Better than linear models for heterogeneous feature integration, but still depends on feature engineering
&
Moderate training cost and efficient inference
&
Feature importance and decision paths provide interpretable evidence
&
Stronger than linear classifiers, but limited on unseen domains
\\

XGBoost
&
\cite{ren2017neural,barbado2019framework}
&
ML-based
&
Gradient boosting decision trees
&
TF-IDF, linguistic indicators, sentiment features, reviewer behavior, rating and temporal metadata
&
Handles heterogeneous features but depends on feature construction
&
Higher training cost than RF, efficient inference
&
Feature importance and SHAP analysis
&
Strong structured-signal modeling but limited transferability
\\

CNN
&
\cite{ren2017neural,barbado2019framework,10652819}
&
DL-based
&
End-to-end convolutional representation learning
&
Word embeddings and automatically learned local semantic patterns
&
More adaptable than handcrafted ML features but sensitive to vocabulary shift
&
Moderate training cost with GPU acceleration
&
Convolution filters reveal important phrases
&
Better linguistic modeling but vulnerable to domain shift
\\

RNN
&
\cite{ren2017neural,monisha2024detection,yao2017automated}
&
DL-based
&
Sequential representation learning
&
Word embeddings with contextual sequential representations
&
Flexible but sensitive to domain-specific distributions
&
Higher training cost due to sequential computation
&
Hidden states encode contextual information
&
Captures contextual deception but limited by long dependency
\\

LSTM
&
\cite{xu2024hybridizeda,wang2018detecting,computers13100264}
&
DL-based
&
Gated sequential representation learning
&
Word embeddings and hidden states capturing long-term dependencies
&
Better than RNN through memory mechanism
&
Higher computational cost than CNN/RNN
&
Hidden states and attention weights provide partial interpretation
&
Strong contextual modeling but vulnerable to domain shift
\\

GNN
&
\cite{duma2025analysis,tang2020generating,shawon2023bengali,Cheng2024FakeReviewerGNN}
&
DL-based
&
Graph representation learning through neighborhood aggregation
&
User-review-product graph embeddings integrating text and behavioral signals
&
Strong relational transfer but depends on graph availability
&
High complexity due to graph construction and message passing
&
Graph structures and influential neighbors provide explanations
&
Strong for coordinated spam detection
\\

Attention Networks
&
\cite{salminen2025decoding,wang2017detecting}
&
DL-based
&
Attention-enhanced neural representation learning
&
Contextual representations weighted by attention scores
&
Improves transferable semantic cues
&
Higher than CNN/RNN due to attention computation
&
Attention weights highlight important features
&
Improves robustness but depends on attention quality
\\

BERT
&
\cite{catelli2023new,10.1145/3497701.3497714}
&
PLM-based
&
Pretraining followed by fine-tuning
&
Transformer contextual embeddings
&
Strong transfer ability through pretrained knowledge
&
High training cost
&
Attention distributions and token attribution
&
Strong cross-domain capability with adaptation
\\

RoBERTa
&
\cite{liu2023data,10.1145/3503162.3503169}
&
PLM-based
&
Transformer pretraining and fine-tuning
&
Deep contextual semantic representations
&
Strong transfer ability
&
Higher cost due to larger parameters
&
Attention and contrastive representations
&
Strong cross-domain performance
\\

DeBERTa
&
\cite{NARAYAN2025100285,prasetyaningrum2025smart}
&
PLM-based
&
Disentangled attention based pretraining
&
Semantic and positional contextual representations
&
Strong adaptation capability
&
Higher computational requirement
&
Attention attribution analysis
&
Strong generalization under adaptation
\\

GPT-series
&
\cite{ignat2024maideupmultilingualdeceptiondetection,Wang2026LLMIC,HUANG2026114529}
&
LLM-based
&
Zero/few-shot prompting and instruction-based discrimination
&
Reasoning-oriented semantic representations
&
Strong prompt-based adaptation
&
Very high inference cost
&
Natural-language explanations but hallucination risk
&
Strong cross-domain transfer but prompt-sensitive
\\

LLaMA-series
&
\cite{FAN2025101521,miah2025hiddenplainsightevaluation}
&
LLM-based
&
Instruction tuning and parameter-efficient adaptation
&
Large-scale causal language representations
&
Flexible through LoRA and continued pretraining
&
High training cost but efficient local deployment
&
Generated rationales with opaque internal states
&
Strong but depends on alignment quality
\\

Instruction-tuned LLMs
&
\cite{bader2023detecting,ignat2024maideupmultilingualdeceptiondetection,Wang2026LLMIC}
&
LLM-based
&
Instruction following and reasoning-based detection
&
Instruction-aligned semantic representations
&
Few-shot adaptation through prompts
&
High latency and computational demand
&
Human-readable explanations but hallucination risks
&
Strong zero/few-shot generalization
\\

\bottomrule

\end{tabular}

\end{adjustbox}

\end{sidewaystable*}

ML methods remain effective when review datasets are relatively small and explicit deception indicators are available.
Models such as SVM, logistic regression, random forest, and gradient boosting methods achieve competitive performance by exploiting handcrafted linguistic, behavioral, and metadata features.
Their major advantages are low computational cost, fast inference, and relatively strong interpretability.
However, their effectiveness is highly dependent on feature engineering, making them vulnerable to evolving deception strategies and cross-domain distribution shifts in increasingly dynamic online environments.

DL approaches alleviate the dependence on manual feature engineering by learning hierarchical representations directly from review texts.
CNN, RNN, LSTM, attention networks, and GNN-based models have demonstrated stronger capability in capturing nonlinear linguistic patterns and user-review interactions.
Nevertheless, these methods usually require larger labeled datasets and considerable computational resources.
In addition, their generalization ability remains limited when transferred to unseen domains because representations are learned from task-specific data.

PLM-based methods represent a major advancement by introducing large-scale linguistic knowledge into fake review detection.
Models including BERT, RoBERTa, and DeBERTa achieve stronger semantic understanding and cross-domain adaptability through pretrained contextual representations and domain-specific fine-tuning.
Therefore, PLM-based methods provide a favorable balance between detection performance and deployment feasibility, making them suitable for practical online review moderation systems with improved scalability and reliability.
However, fine-tuning still requires domain-specific labeled data, and computational costs remain higher than conventional neural models.

LLM-based methods further extend fake review detection from classification toward reasoning-oriented deception analysis.
GPT-series, LLaMA-based models, and instruction-tuned LLMs can perform zero-shot and few-shot detection through prompting, requiring less task-specific annotation.
These models can identify AI-generated reviews, implicit deception patterns, and cross-domain fraudulent behaviors.
However, their deployment is still restricted by high inference cost, latency, privacy concerns, and potential hallucinated explanations.

The choice of detection paradigm should be determined by the trade-off between accuracy requirements, computational resources, and deployment objectives.
ML-based approaches remain preferable for lightweight monitoring scenarios with limited data and strict interpretability requirements.
DL-based models are suitable when large-scale labeled datasets are available and complex linguistic patterns need to be captured.
PLM-based approaches currently provide the most practical solution for general-purpose fake review detection because they achieve strong semantic understanding with manageable deployment costs and improved adaptability.
In contrast, LLM-based approaches are useful for emerging AI-generated fake reviews and scenarios requiring adaptive reasoning, although additional optimization is needed for large-scale real-time deployment.
Future detection frameworks are likely to evolve toward hybrid architectures that integrate the efficiency of traditional models, the representational power of PLMs, and the reasoning capabilities of LLMs.

\subsection{Fusion-Oriented Practical Insights}
Recent advances in LLMs and multimodal foundation models have substantially changed both fake review generation and detection. Traditional fake review detection mainly focused on manually fabricated deceptive opinions through linguistic patterns, sentiment inconsistencies, and reviewer behavioral signals. Recent fake reviews can be fluent, adaptive, and context-aware. Detection research is therefore shifting from text classification toward generation-aware, multimodal, and knowledge-enhanced fusion frameworks. In this section, fusion is interpreted as the operational combination of evidence sources at feature, representation, graph, or decision levels. This framing distinguishes methods that simply add more features from methods that model cross-source interactions, source reliability, and uncertainty under missing or noisy signals. It also further clarifies why fake review detection fits the scope of information fusion research rather than only text classification alone.

The emergence of large language models has lowered the cost of generating deceptive reviews.
Unlike traditional fake reviews written by human spammers, LLM-generated reviews can achieve high grammatical quality, semantic coherence, and emotional consistency, making them difficult to distinguish from authentic user-generated content.
Recent studies have demonstrated that models such as GPT-series can generate realistic deceptive hotel and product reviews across different domains and languages.

For example, \citet{ignat2024maideupmultilingualdeceptiondetection} introduced MAiDE-up, a multilingual deception detection benchmark containing GPT-generated hotel reviews. Their findings showed that AI-generated reviews exhibit increasingly human-like characteristics and pose significant challenges to existing detection models. This further indicates that future detection systems cannot rely solely on lexical or stylistic features.

Beyond simple text generation, instruction-tuned LLMs enable controllable fake review generation by incorporating additional constraints, including product characteristics, user profiles, sentiment polarity, and writing styles.
Such controllability allows attackers to generate diverse and personalized deceptive reviews while reducing repetitive patterns commonly exploited by traditional detectors.
Therefore, fake review detection has evolved from identifying abnormal texts toward modeling the interaction between generated content, user behavior, and contextual evidence.

Recent studies have explored the use of LLMs themselves as detection tools for advanced fake review analysis tasks.
\citet{Wang2026LLMIC} proposed an LLM-based fake review detection framework from the perspective of implicit characteristics, showing that LLMs can capture semantic inconsistencies and hidden deception patterns beyond surface-level linguistic signals.
Similarly, \citet{HUANG2026114529} investigated verbal deception patterns in LLM-generated reviews and revealed that generated texts may contain distinctive discourse structures and reasoning patterns.

Another important trend is the emergence of adversarial generation-detection cycles.
Generative models can be used to simulate future attack strategies and produce challenging training samples for detector improvement.
For instance, \citet{FAN2025101521} proposed an LLM-enhanced deep learning framework for fake review detection, demonstrating that generated adversarial samples can improve model robustness against evolving deception strategies.

Nevertheless, several challenges remain.
First, the rapid evolution of generative models creates a continuously changing detection environment.
Second, prompt engineering and style imitation techniques allow attackers to intentionally remove detectable generation artifacts.
Third, hybrid human-AI generated reviews introduce additional ambiguity, making the boundary between authentic and artificial content increasingly unclear.
Future research should investigate generation-aware detection mechanisms that jointly model linguistic evidence, generation probability, behavioral consistency, and external knowledge.

Although early fake review detection studies focused on textual content, online reviews increasingly contain heterogeneous information sources, including images, videos, ratings, user profiles, and social interactions.
The availability of multimodal foundation models has motivated a transition from text-centric detection toward multimodal authenticity verification.
Images provide important complementary evidence for identifying deceptive reviews.
For example, fraudulent reviewers may reuse product images, upload unrelated photos, or generate synthetic visual content to strengthen misleading claims.
Therefore, integrating visual information with textual semantics can provide additional signals that cannot be captured by text-only models.

Recent multimodal detection frameworks typically combine textual encoders, visual representation models, and behavioral features.
Text representations extracted by PLMs or LLMs are integrated with image embeddings from vision models such as CLIP-based architectures to capture cross-modal consistency.
A genuine review generally exhibits semantic alignment between textual descriptions and attached images, whereas deceptive reviews may contain inconsistencies across modalities.
The emergence of multimodal large language models, such as GPT-4V and other vision-language models, may further expand fake review detection when reviews contain both text and images.
Direct evidence in fake review benchmarks remains limited, but these models can potentially support joint interpretation of text, images, and contextual information.
We did not identify a video-specific fake review benchmark in the coded corpus, so video-based detection is treated as an open research problem rather than an established method.
However, practical deployment remains challenging due to high computational costs, latency constraints, and limited interpretability of multimodal reasoning processes.

Another challenge is the lack of large-scale multimodal fake review benchmarks with realistic and diverse cross-platform characteristics.
Existing datasets are predominantly text-based, while multimodal datasets containing synchronized reviews, images, user behaviors, and temporal information remain limited.
Future research should focus on constructing multimodal benchmarks and developing efficient multimodal detection architectures that balance accuracy, robustness, and deployment efficiency.

For practical deployment, fusion design should match the available evidence sources.
Text-only systems remain useful when behavioral or graph data are unavailable.
Text-behavior fusion is more appropriate for platforms with reliable user histories and rating timelines.
Graph-based fusion is useful for detecting coordinated manipulation, while multimodal fusion is needed when reviews include images or videos.
These choices also affect privacy, computational cost, interpretability, and robustness under missing sources.

\section{Technological Challenges and Research Trends}
\label{sec:challenges}

The preceding sections have traced fake review detection from traditional machine learning through deep learning to PLM- and LLM-based methods.
This evolution has improved reported performance, but it has also exposed new weaknesses in evidence reliability, benchmark validity, and multi-source fusion.
The remaining technical agenda centers on unresolved challenges in LLM-era fake review detection and the research directions needed to address them.

\subsection{Challenges}
\textbf{AI-generated deceptive reviews.}
The evolution of LLMs such as the GPT series, ChatGPT, and LLaMA has made AI-generated fake reviews difficult to distinguish from genuine ones.
Fabricated reviews produced by these models can exhibit natural syntax, coherent semantics, and human-like style~\cite{brown2020language}.
Detection systems that rely on surface linguistic features are vulnerable when attackers use prompt engineering, style imitation, or paraphrasing to suppress detectable artifacts.
Models therefore need to reason about semantic consistency, behavioral plausibility, and source reliability under changing attack strategies.

\textbf{Cross-domain and cross-platform generalization.}
The effectiveness of PLMs and LLMs depends heavily on the availability of large and diverse training corpora, yet fake reviews vary across platforms, domains, languages, and cultural contexts.
Domain-adaptive pretraining can reduce part of this gap~\cite{gururangan2020don}, but many PLM-based detectors still degrade when transferred to new review domains.
Cross-domain generalization cannot be solved by model architecture alone.
It requires source-wise shift analysis, domain adaptation, and evaluation protocols that explicitly test transfer across product categories, platforms, and languages.

\textbf{Unreliable, missing, and conflicting evidence sources.}
Beyond textual content, timestamps, rating patterns, user histories, product metadata, images, and graph relations carry important signals for review authenticity.
These sources are often incomplete, noisy, private, or adversarially manipulated.
\citet{Xu2025Interpretable} proposed a multimodal approach that combines review text with user behavioral data and rating information, showing that fusion can detect complex deception that eludes text-only detectors.
The remaining challenge is to design fusion mechanisms that remain reliable when one source is missing, sparse, or inconsistent with other sources.

\textbf{Interpretability and decision accountability.}
Fake review detection systems are used in moderation and reputation management settings where false accusations can harm users and merchants.
Text-only explanations are often insufficient because the final decision may depend on behavior, graph structure, temporal evidence, or multimodal consistency.
Fusion-based detectors need source-level explanations that show which evidence sources contributed to the prediction, how conflicts were resolved, and how uncertain the final decision is.

\subsection{Future Directions}

\paragraph{Controllable data augmentation and adversarial robustness.}

When fake review samples are costly to obtain and class distributions are imbalanced, data augmentation can improve detector training.
\citet{liu2025data} proposed using generative models such as GPT and GLM to synthesize fake review samples for training set expansion.
Future work should control emotional intensity, semantic coherence, rating behavior, and temporal patterns in generated samples so that synthetic data better reflects realistic attack scenarios.
Adversarial training can also simulate malicious generation strategies and improve robustness against both human-written and AI-generated reviews~\cite{qian2025glm}.

\paragraph{Cross-domain adaptation and multi-task learning.}

The challenge of cross-domain generalization is tied to distribution shift across platforms and domains.
Approaches that combine domain-adaptive pretraining with multi-task learning can share semantic and affective knowledge across related tasks such as fake review detection, sentiment analysis, and topic identification.
Future work should investigate LLM-based representation learning frameworks that jointly train on multi-task and multi-domain data~\cite{catelli2023new}.
Domain discriminators, adaptive modules, and source-wise alignment objectives can further reduce degradation between source and target domains.

\paragraph{Uncertainty-aware and missing-source fusion.}

Fusion-based detectors should not assume that all evidence sources are available and reliable.
Future systems need uncertainty-aware fusion mechanisms that estimate source reliability, calibrate prediction confidence, and degrade gracefully when metadata, behavior histories, graph links, or visual evidence are missing.
Source-wise ablation, confidence calibration, and missing-modality stress tests should become standard evaluation components.

\paragraph{Explainability enhancement and hallucination mitigation.}

Although LLMs support semantic reasoning in fake review detection, their opaque internal representations and potential for hallucinated explanations constrain deployment in moderation scenarios.
Future research should emphasize explanations that identify the evidence behind each decision, including linguistic inconsistencies, behavioral anomalies, graph-based signals, and multimodal conflicts~\cite{zhang2025siren}.
Reasoning paths generated by LLMs through chain-of-thought prompting or other explanatory strategies may contain inaccurate justifications~\cite{wei2022chain,huang2025survey}.
Explainability methods therefore need verification mechanisms and confidence estimation to prevent erroneous explanations from misleading human reviewers and automated decision processes~\cite{10.7717/peerj-cs.2345}.

These directions also differ in urgency. Benchmark auditing, source-wise ablation, and confidence calibration are immediate priorities because they support reliable comparison of existing methods. Cross-domain adaptation and missing-source fusion form the next stage, while continual defense against adaptive generators and accountable human oversight require longer-term study.

Taken together, these challenges point to a broader change in how fake review detection should be conceptualized.
Detection is no longer only a text classification problem.
It is a multi-source inference task that draws on textual, behavioral, structural, temporal, visual, and contextual evidence.
Future progress depends on fusion mechanisms that remain reliable under adversarial generation, domain shift, missing sources, and uncertain explanations.

\section{Conclusion}
\label{sec:conclusion}

This survey reviewed fake review detection from rule-based and feature-engineering approaches through deep learning and PLMs to LLM-based methods.
Across this trajectory, the central challenge has shifted from identifying surface linguistic anomalies toward reasoning about semantic deception, manipulative intent, and evidence reliability.
PLM-based methods improve contextual semantic modeling, but they remain vulnerable to adversarial generation, cross-domain distribution shift, and limited interpretability.
LLM-based approaches expand the detection toolkit through prompt-driven reasoning, semantic representation learning, generative discriminative modeling, and data augmentation.
At the same time, LLMs intensify the threat by enabling fluent and contextually coherent fabricated reviews.
The main finding of this survey is that fake review detection is becoming a multi-source information fusion problem rather than a text classification problem alone.
Future detectors need to integrate textual semantics, reviewer behavior, temporal metadata, relational graph structures, multimodal content, and external knowledge while accounting for source reliability and uncertainty.
Progress in this direction is necessary for reliable review moderation and for preserving the consumer information value of online reviews.

\printcredits

\section*{Acknowledgements}

This work was supported by the Central Guidance Fund for Local Science and Technology Development
(Grant No. \texttt{QKHZYD[2025]035}), and the Guizhou Provincial Major Scientific and Technological Program
(Grant No. \texttt{QKHCG[2024]ZD018}).

\bibliographystyle{cas-model2-names-unsrt}
\bibliography{references}

\end{document}